%% file: main.tex
\documentclass[runningheads]{llncs}

\usepackage[mobile]{eccv}
\input{preamble}

\usepackage{eccvabbrv}

\usepackage{graphicx}
\usepackage{booktabs}
\usepackage{multirow}

\usepackage[accsupp]{axessibility}  

\usepackage{hyperref}

\usepackage{orcidlink}

\begin{document}

\title{Map2World: Segment Map Conditioned \\ Text to 3D World Generation}


\author{Jaeyoung Chung\textsuperscript{*}\inst{1} \qquad
Suyoung Lee\textsuperscript{*}\textsuperscript{\textdagger}\inst{1} \qquad
Jianfeng Xiang\inst{2} \\
Jiaolong Yang\inst{2} \qquad
Kyoung Mu Lee\inst{1}}

\authorrunning{J. Chung et al.}

\institute{Seoul National University, \email{\{robot0321,esw0116,kyoungmu\}@snu.ac.kr} \and 
Microsoft Research Asia, \email{\{t-jxiang,jiaoyan\}@microsoft.com} \\
}

\maketitle
\renewcommand{\thefootnote}{*}
\footnotetext{indicates equal contribution.}
\renewcommand{\thefootnote}{\arabic{footnote}}
\renewcommand{\thefootnote}{\textdagger}
\footnotetext{works done during internship at Microsoft Research.}
\renewcommand{\thefootnote}{\arabic{footnote}}

\input{fig/teaser}
\input{sec/0_abstract}

\input{sec/1_intro}
\input{sec/2_relatedwork}

\input{sec/3_method}

\input{sec/4_experiment}

\input{sec/5_conclusion}

%
%
\bibliographystyle{splncs04}
\bibliography{main}

\newpage
\input{sec/X_supple}

\end{document}

%% file: preamble.tex
\usepackage{url}            
\usepackage{booktabs}       
\usepackage{amsfonts}       
\usepackage{amsmath}       %
\usepackage{nicefrac}       
\usepackage{microtype}      
\usepackage{xcolor}         
\usepackage{gensymb}
\usepackage{amssymb}
\usepackage{subcaption}
\usepackage{bm}
\usepackage{bbm}
\usepackage{graphicx}
\usepackage{multirow}
\usepackage{subcaption}



%% file: fig/teaser.tex
\begin{figure}
    \centering
    \includegraphics[width=1\linewidth]{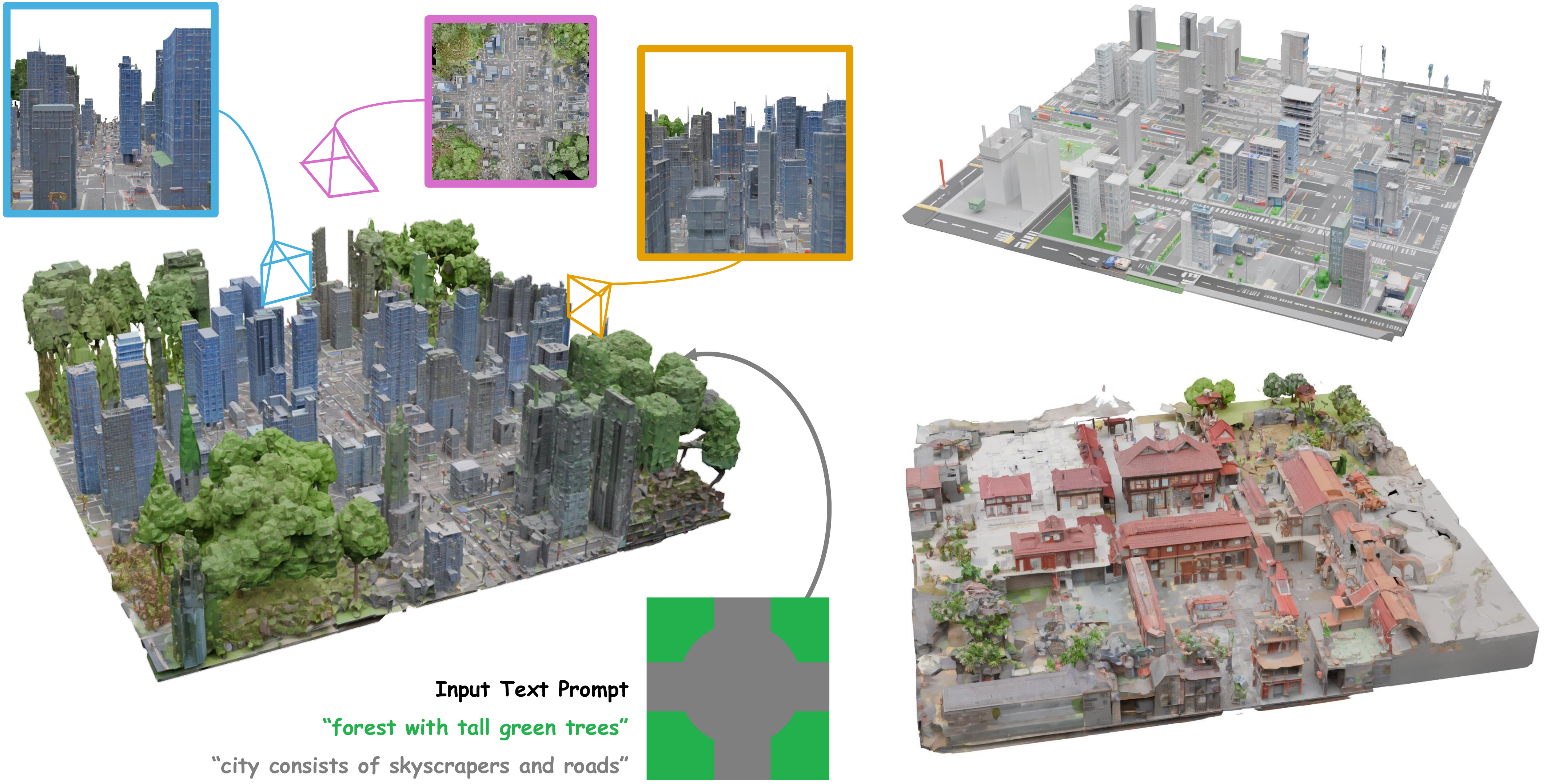}
    \captionof{figure}{
        (Left) A generated sample of the center of the city, where the four edges consist of green forest.
        Our model takes as input a segment map with text prompts for each segment and produces a world that corresponds to the input segment map (violet box) with high quality (blue and orange boxes).
        (Right) Two examples of generated worlds in large-scale.
    }
    \label{fig:teaser}
\end{figure}
\vspace{-8mm}

%% file: sec/0_abstract.tex
\begin{abstract} \label{sec:abstract}
3D world generation is essential for applications such as immersive content creation or autonomous driving simulation.
Recent advances in 3D world generation have shown promising results; however, these methods are constrained by grid layouts and suffer from inconsistencies in object scale throughout the entire world.
In this work, we introduce a novel framework, Map2World, that first enables 3D world generation conditioned on user-defined segment maps of arbitrary shapes and scales, ensuring global-scale consistency and flexibility across expansive environments.
To further enhance the quality, we propose a detail enhancer network that generates fine details of the world.
The detail enhancer enables the addition of fine-grained details without compromising overall scene coherence by incorporating global structure information.
We design the entire pipeline to leverage strong priors from asset generators, achieving robust generalization across diverse domains, even under limited training data for scene generation.
Extensive experiments demonstrate that our method significantly outperforms existing approaches in user-controllability, scale consistency, and content coherence, enabling users to generate 3D worlds under more complex conditions.
\keywords{3D World Generation \and Structured Latent \and 3DGS}
\end{abstract}

%% file: sec/1_intro.tex
\section{Introduction} \label{sec:intro}
Three-dimensional (3D) world generation plays a pivotal role in a wide range of industrial applications, including immersive content creation, gaming, autonomous driving simulation, and virtual realities. 
While recent years have witnessed remarkable progress in 3D asset or object generation, extending these capabilities to the world scale remains challenging. 
The primary bottleneck is the lack of high-quality world-level datasets, which are far more difficult to construct than object-centric collections. 
Several works, such as BlockFusion~\cite{wu2024blockfusion}, NuiScene~\cite{lee2025nuiscene}, LT3SD~\cite{meng2025lt3sd}, WorldGrow~\cite{worldgrow2025} or SCube~\cite{ren2024scube}, trained their own generator with existing datasets. 
However, these methods can only generate scenes within a limited domain, such as indoor or driving scenes, severely limiting their applicability.

To circumvent the absence of world-scale data, researchers often resort to leveraging the power of pre-trained diffusion models.
One approach combines image diffusion models with a depth estimator to lift generated 2D images into 3D; however, these pipelines suffer from view-dependent inconsistencies, yielding incomplete reconstructions.
Another direction employs video diffusion models to generate scene-like sequences.
Still, these methods struggle with 3D-consistency and are fundamentally constrained by the limited memory span of diffusion models.

Recent 3D generation models, such as TRELLIS~\cite{xiang2025structured} and CLAY~\cite{zhang2024clay}, demonstrate the feasibility of high-quality, domain-agnostic 3D generation at an asset scale.
Building on these foundations, several attempts have been made to scale asset generators to the world level.
For instance, SynCity~\cite{engstler2025syncity} divides the ground into grids and generates proper 3D assets for each tile using TRELLIS. Then, the asset boundaries are blended using an image inpainting module.
Although these approaches effectively leverage the expressive power of 3D asset generators, the failure to model relationships among generated assets poses challenges for large-world generation.
For example, contextual disconnectedness between adjacent assets and inconsistencies in object scale can make the overall scene appear less coherent and harmonious.
In addition, the assets should be arranged in a grid-like manner, which is impractical in real scenarios, as district boundaries are irregularly shaped.

In this paper, we present a novel text-conditioned 3D world generation framework called \textbf{Map2World} that builds on TRELLIS and explicitly addresses the limitations.
To ensure global context and scale consistency, we introduce a multi-diffusion strategy applied within the structured latent space.
By coordinating overlapping diffusion windows, our method not only preserves the latent prior of TRELLIS but also enables seamless connections that extend beyond the boundaries of individual cubes. 
This design naturally supports arbitrary resolution, as the scene can be generated progressively while maintaining coherence across local neighborhoods.
Furthermore, our approach can flexibly incorporate semantic maps as conditions without additional training, enabling controllable, semantically aligned scene synthesis.
The ability to balance local detail, global context, and semantic structure allows our method to generate globally coherent, high-resolution 3D scenes that surpass the limitations of existing modular approaches.
Through extensive experiments, we demonstrate that our framework significantly improves both structural fidelity and perceptual realism, paving the way toward scalable and versatile 3D scene generation.

Our contributions can be summarized as follows:
\begin{itemize}
    \item Flexible segment map conditioning: generates 3D worlds from any type of user-defined segment maps, not limited to grid-based layouts.
    \item Consistent detail enhancement: adds fine details to the assets while preserving overall structure.
    \item Domain-generalized world generation: leverages powerful asset generator priors to achieve robust generation across domains with limited data.
\end{itemize}





%


%% file: sec/2_relatedwork.tex
\section{Related work} \label{sec:relatedwork}

\noindent\textbf{3D world generation.}
Compared to 3D asset generation, creating a world is considered a more complex task, as a world consists of multiple objects that are harmonized with each other.
We classify the approaches to generate a 3D world into two categories: 3D reconstruction from generated rendered views, and direct explicit 3D generation.

The former approach uses diffusion models to generate the images or videos, and reconstructs a 3D scene from the generated views.
2D image diffusion model-based approaches~\cite{hoellein2023text2room,chung2023luciddreamer,yu2024wonderjourney,yu2025wonderworld,li2024dreamscene,shriram2024realmdreamer} lift the pixels of the initial image into 3D with a monocular depth estimator. 
The image is then outpainted, and the generated region is lifted and stitched to the original scene to expand the world.
The video diffusion model-based approaches~\cite{wang2025videoscene,zhang2025world,liu20243dgs,yan2025streetcrafter} generate a video that navigates a virtual environment and reconstructs the scene from the video frames. 
Yet, these methods cannot guarantee the 3D consistency since the diffusion models are not trained to be aware of such consistency.

The other approach directly creates the scene with explicit 3D representations, achieving perfect 3D consistency.
Some methods, such as BlockFusion~\cite{wu2024blockfusion}, NuiScene~\cite{lee2025nuiscene}, and LT3SD~\cite{meng2025lt3sd}, train the generator from scratch to learn the distribution of cubes that correspond to a scene part.
However, they share two major drawbacks resulting from the limited availability of scene datasets.
First, these models can only generate geometry, and the generated scene does not have textures.
Second, the domain of the generation is limited to the dataset they used for training.
SCube~\cite{ren2024scube} and InfiniCube~\cite{lu2024infinicube} add color estimation and use sparse-voxel-hierarchy~\cite{ren2024xcube} to improve global consistency, but they are also constrained to representing only driving scenes, which poses a significant limitation.
Several works~\cite{engstler2025syncity,zheng2025constructing} leverage off-the-shelf 3D asset generators~\cite{xiang2025structured,ren2024xcube,lin2023magic3d,chen2023fantasia3d,zhang2024clay} and expand the scope of generation to 3D world.
Although the quality of each asset can be prominent, the overall quality of the scene is highly dependent on the dataset.
The strong prior knowledge of a high-quality 3D asset generator enhances both the quality and diversity of generated results; however, the spatial resolution of the generator’s output is too limited to represent an entire world.
SynCity addresses this issue by dividing the space into multiple grid tiles, generating assets for each tile, and then merging them.
Nevertheless, this approach suffers from weak connectivity between tiles and is unable to create large objects with arbitrary shapes.
In contrast, our model supports seamless world generation from arbitrary segment maps using the latent fusion strategy.

\noindent\textbf{Scaling diffusion models to large spatial extents.}
The idea of large-scale 3D generation can originate from the 2D diffusion literature, where training-free techniques have been developed to expand the resolution of pretrained models beyond their limits.
Patch-based approaches synthesize large images by denoising overlapping patches and stitching them together, effectively bypassing the base model’s resolution and memory constraints~\cite{ding2023patched, jiang2025latent}.
However, these methods typically rely on rigid grid tilings and heuristic blending, which can introduce boundary artifacts and offer limited flexibility for region-wise semantic control. 
Outpainting-based methods progressively extend an input canvas beyond its original field of view, enabling directional or large-aspect-ratio image expansion~\cite{kim2021painting, wu2023panodiffusion, chen2024follow, song2025progressive}.
Despite this flexibility, they remain confined to 2D planar domains and are usually driven by a single global context, providing only coarse control over the semantics of newly generated regions. 
Multi-window diffusion frameworks treat a large canvas as a collection of overlapping windows that are denoised jointly, allowing different prompts or conditions to be assigned to different spatial regions while maintaining global coherence through latent or score fusion~\cite{bar2023multidiffusion, jimenez2023mixture, du2024demofusion, lee2023syncdiffusion, lee2024streammultidiffusion}.
Building on this idea, we extend the multi-window paradigm to volumetric 3D, operating on arbitrary-shaped, user-defined 3D regions and fusing their denoising trajectories in a shared 3D latent space to construct large, coherent 3D scenes.


%% file: sec/3_method.tex
\input{fig/method}

\section{Preliminary: Structured Latent and Generation Pipeline} \label{sec:method1}

We summarize TRELLIS~\cite{xiang2025structured}, a state-of-the-art 3D asset generation model, using the new representation, structured latent, and the two-stage generation process.
The structured latent (or SLAT)~\cite{xiang2025structured} encodes geometry and appearance with a set of local latents on a 3D grid,
\begin{equation}
    \bm{s} = \left\{\left(\bm{z}_i, \bm{p}_i\right)\right\}^L_{i=1},  
\end{equation}
where $\bm{p}_i \in \{0,1,\ldots,N\text{-}1\}^3$ is the positional index of an active voxel in the $N^3$ 3D grid, $\bm{z}_i \in \mathbb{R}^{C}$ denotes a latent vector that encodes geometry and appearance at the corresponding position $\bm{p}_i$, and $L$ is the number of active voxels.

The structured latent is generated by a two-stage generation pipeline, from geometry to texture, based on the rectified flow model.
Here, we assume the input text prompt $y$ is given as a condition.
For the first stage, they generate sparse structure $\{\bm{p}_i\}^L_{i=1}$ by iteratively applying the flow Transformer $\bm{\mathcal{G}}_S$ conditioned by $y$ to the Gaussian noise.
The fully denoised sparse structure latent is passed through the decoder, denoted as $\bm{\mathcal{D}}_S$, and transformed into a signed 3D scalar field where the voxels with positive values are filled with contents and called active.
We record a set of position indices of all active voxels as $\{\bm{p}_i\}^L_{i=1}$.
In the second stage, the set of latent features $\{\bm{z}_i\}^L_{i=1}$ is also estimated through iterative denoising steps using the Transformer called $\bm{\mathcal{G}}_L$, similar to the first stage.
The structured latent can be transformed with various types of 3D representations, including 3D Gaussian splatting~\cite{kerbl20233d}, radiance fields~\cite{gao2023strivec}, and mesh~\cite{shen2023flexible}, by putting the latent to the corresponding decoder ($\bm{\mathcal{D}}_L$).
In this experiment, we mainly focus on generating 3DGS as a final 3D representation.

\section{Proposed Method} \label{sec:method}

We describe how Map2World exploits the knowledge of a 3D asset generator to synthesize a 3D world.
In \cref{sec:method2}, we present our latent fusion strategy to expand generation to a wider scene and to support conditioning on multiple text-driven spatial regions of arbitrary shapes.
\cref{sec:method3} proposes a detail enhancing network to enrich the details of the world by manipulating the structured latent.
Here, we explain the design choice of the detail enhancer to leverage the representational power of TRELLIS while preserving the content of the input large 3D scene.
Finally, \cref{sec:method4} introduces decoder fine-tuning to generate high-quality 3D scenes.
The overall generation process is illustrated in \cref{fig:method}.

\subsection{Expanding Spatial Regions in 3D Latent Space} \label{sec:method2}

The active voxel space of assets generated by the pre-trained TRELLIS model is limited to a cube of size 64.
While this resolution may be sufficient for representing a single object, it is inadequate for modeling a large world composed of numerous objects.
To achieve spatial expansion using the pre-trained model, we draw inspiration from spatial expansion techniques in 2D diffusion~\cite{bar2023multidiffusion} and apply the idea to both 3D volumetric tensors and sparse structures in the rectified-flow Transformers, $\bm{\mathcal{G}}_S$ and $\bm{\mathcal{G}}_L$.
Our method supports precise and controllable world generation conditioned on fine-grained, pixel-level spatial regions annotated with textual conditions.
We further describe a 3D noise initialization strategy that reinforces global scale consistency during progressive generation.

\subsubsection{Latent fusion for rectified flow models.}
We conceptually split the space into a set of overlapping 3D cube windows $\{\Omega_j\}$, where each window spans a cubic region of size 64 and adjacent windows overlap by half of their spatial resolution.
From a given position $\mathbf{x}$, we update its local latent by aggregating the velocity field predictions $v_j(\mathbf{x})$ from all windows spatially covered as $\mathcal{A}(\mathbf{x}) = \{ \, j \mid \mathbf{x}\in\Omega_j \,\}$. Using a shared 3D Gaussian kernel $W(\cdot)$, the fused velocity for the position $\mathbf{x}$ is: 
\begin{align}
v_t(\mathbf{x}|y) = \frac{
\sum_{j\in\mathcal{A}(\mathbf{x})} W(\mathbf{x}-\mathbf{c}_j)\, v_{t,j}(\mathbf{x}|y)
}
{
\sum_{j\in\mathcal{A}(\mathbf{x})} W(\mathbf{x}-\mathbf{c}_j)
},
\end{align}
where $\mathbf{c}_j$ is the center of $\Omega_j$.
This strategy is applied to both $\bm{\mathcal{G}}_{S}$ and $\bm{\mathcal{G}}_{L}$, and the latent at $\mathbf{x}$ is updated at every step according to the rectified-flow formulation:
\begin{align}
    \bm{s}_{t-1}(\mathbf{x}|y) &= \bm{s}_t(\mathbf{x}|y) - \Delta t \cdot v_t(\mathbf{x}|y).
\end{align}

\subsubsection{Segment-map-guided latent fusion.}
Map2World supports 3D world generation conditioned by multiple text prompts with a segment map.
Assuming that we have $K$ labels, we define $M_k$ as a binary map that indicates the region of the $k$-th label, and $y_k$ as the corresponding text prompt.
Here, the velocity at the position $\mathbf{x}$ is calculated by the weighted sum of velocities estimated for each label:
\begin{equation}
\tilde{v}_t(\mathbf{x}) = \frac{
\sum_{k=1}^{K} \left(M_k(\mathbf{x}) \odot G(\sigma_t)\right) \cdot v_t(\mathbf{x}|y_k)
}
{
\sum_{k=1}^{K} \left(M_k(\mathbf{x}) \odot G(\sigma_t)\right)
}.
\end{equation}

$G(\sigma_t)$ is a normalized 3D Gaussian kernel with zero mean and standard deviation $\sigma_t$ to ensure smooth transition across regions, improving denoising stability.
Here, $\sigma_t$ is controlled by the diffusion time step, starting from a large value to produce soft boundaries and gradually changing toward a sharp mask as $t$ decreases.
The smooth transitions of the segment mask across regions improve stability during the diffusion process.
It is noteworthy that this pipeline can be applied to any shapes of $M_k$, whereas the existing framework, SynCity, can only generate objects of the same type in a square-shaped region.

\subsubsection{Optimization for scale-aware initial latent.} \label{sec:optinit}
We observe two consistent behaviors in the noisy latent space. First, coarse geometric structures vary significantly with different initial noise samples. Second, similar initial noise samples tend to produce similar outputs. These observations indicate that the mapping from the initial latent $x_T$ to the final generation is locally smooth yet highly sensitive to the initial condition. 

Although TRELLIS does not explicitly split the scale of scene, we empirically find that sparse structures exhibit different scene scales, and structures with similar scales tend to appear near each other in the latent space, as illustrated in \cref{fig:ablation_init}. This suggests that valid sparse structures lie on a scale-dependent manifold. 
To steer generation toward a desired scale, we optimize the initial noisy latent inspired by the idea of initial noise optimization~\cite{baek2025sonic}.
For the sparse structure $S_T$ and rectified flow models $\mathcal{G}_S$, we approximate the denoising trajectory by
\begin{equation}
S(t) \approx S_T + (1 - \tfrac{t}{T})[\mathcal{G}_S(S_T)-S_T]_{\mathrm{sg}},
\end{equation}
to calculate the gradient without backpropagating through the full denoising trajectory. Here, $[\cdot]_{\mathrm{sg}}$ denotes the stop-gradient operator. Under this approximation, the initialization is optimized via
\begin{equation}
\mathcal{L}_{\text{linear}} = \big\|y - \mathcal{M}\!\left([\mathcal{G}_L(S_T)-S_T]_{\mathrm{sg}} + S_T\right)\big\|_2^2 ,
\end{equation}
where $\mathcal{M}$ denotes the target mask and $y$ represents the target constraint for guiding the scale.
Using a representative sparse structure empirically selected from samples, we adjust the scale by defining the excluded regions and the ground as the optimization target $y$.
Further, to stabilize optimization under large learning rates required for early structural updates, we parameterize the sparse structure feature $S$ in the spectral domain using a 3D FFT. This parameterization stabilizes the optimization trajectory and enables the use of a larger learning rate.




\subsection{Enriching Details in 3D Latent Space}
\label{sec:method3}

The generated structured latent represents the geometry and the texture of an entire world.
Since the entire world consists of numerous objects with a ground, it is almost impossible to encode all the details of the world into the latent space with limited capacity.
Thus, we propose a detail-enhancing framework which increases the resolution of the world conditioned by the latent of the input world.
It is challenging to train a network that directly adds details in 3D space since collecting such data pairs for detail enhancement is infeasible.
Instead, we design the detail enhancer to learn the relationship between coarse and fine scenes in the latent space, thereby maximizing the utilization of TRELLIS’ prior knowledge.
To construct dataset pairs for training, we segment existing 3D scene data into cubes. Each cube is designed to contain a sufficient amount of content. Subsequently, each cube is divided into two parts along each axis, resulting in a total of eight smaller cubes.
Both the large cube and the eight small cubes are then passed through the TRELLIS encoder.
Since each latent corresponding to a small cube encodes a relatively smaller spatial region, it contains more detailed information than an input scene latent.
The detail enhancer is trained to estimate the eight latent tensors, where each corresponds to a small cube, conditioned on the latent tensor of the input scene.
While existing auto-regressive pipelines such as BlockFusion~\cite{wu2024blockfusion} and NuiScene~\cite{lee2025nuiscene} aim to predict latents for rendering individual cubes, they face challenges in maintaining global consistency, as these models rely solely on the information of neighboring cubes.
In contrast, our method incorporates a latent variable that encapsulates the entire scene as a condition, and it can achieve high consistency across the whole scene.

We denote the input 3D representation inside a large cube as $\mathcal{C^O}$ and the split small cubes as $\mathcal{C}^j$, where $j=0,1,...,7$.
The structured latent of the large cube is denoted as $\bm{s}^\mathcal{O}=\{(\bm{z}_i^\mathcal{O}, \bm{p}_i^\mathcal{O})\}_i$, and the structured latent of the small cube at index $j$ is denoted as $\bm{s}^j = \{(\bm{z}_i^j, \bm{p}_i^j)\}_i$.


\subsubsection{Network architecture.}
The goal of the detail enhancer is to generate the structured latent of small cubes ($\{\bm{s}^j\}$) conditioned by the structured latent of the input scene ($\bm{s}^\mathcal{O}$).
Instead of estimating the structured latent for all small cubes at once, the detail enhancer receives the cube index ($j$) and predicts the corresponding structured latent($\bm{s}^j$).
When designing the network for detail enhancement, we propose an additional module that accommodates our new conditions and integrates it with the existing flow Transformer of TRELLIS.
The rationale behind this design choice is twofold: first, due to the scarcity of world-level data, it is impractical to develop an entirely new architecture and train all parameters from scratch; second, since the input conditions also reside within TRELLIS’s latent manifold, leveraging the same structure facilitates more efficient processing and integration of condition information.

We employ structured latents from two different cubes as conditioning inputs for detail enhancing network.
First, the structured latent of a large cube, $\bm{s}^\mathcal{O}$, provides rough information about the target cube.
To leverage the information, we extract only the part of the latent corresponding to the spatial location of our target cube, which we denote the truncated latent as $\bm{s}^{\mathcal{O}|j}$.
This structured latent plays a role analogous to a low-resolution image in image super-resolution.
Thus, we concatenate the noise and the condition latent along the channel axis and define an MLP layer ($\bm{F}_\theta$) to mix the information, which is proven effective in a diffusion-based image enhancement field~\cite{saharia2022photorealistic,rombach2022high}.
We denote the channel dimensions of the noise feature and latent feature tensor as $c$ and $C$, respectively; then, the input dimension of the MLP layer is $(c+C)$, and the output dimension is $c$, which is the same as the dimension of the noise feature.

Next, while $\bm{s}^{\mathcal{O}|j}$ can be sufficient to enhance the details of the corresponding small target cube, relying solely on this condition does not guarantee that the geometry and textures between adjacent cubes will be seamlessly connected.
To achieve the seamless connection, we also incorporate the structured latent of adjacent cubes facing the target cube.
The set of structured latent for adjacent cubes with target index $j$ is denoted as $\bm{s}^{Adj(j)}$.
Here, $Adj(j)$ denotes the position indices of cubes that are adjacent to the $j$-th cube.
In order to concatenate the latent features with the noise, as we did with the large cube latent ($\bm{s}^{\mathcal{O}|j}$), we temporarily expand the spatial size of the noise.
The expanded noise is concatenated channel-wise with the adjacent latent and then compressed through the MLP layer.
We note that the MLP layers for $\bm{s}^{\mathcal{O}|j}$ and $\bm{s}^{Adj(j)}$ share the same parameters.
The mixed feature after the MLP layer is then fed into the flow Transformer of the original model($\bm{\mathcal{G}}_{S}$ and $\bm{\mathcal{G}}_{L}$).
From the predicted flow, we crop the expanded region and retain only the portion corresponding to the target cube.
The overall process can be summarized as follows:
\begin{equation}
    \bm{v}_\theta\left(\bm{s}^j, t\right) = \bm{\mathcal{G}}_{S/L}\left( \bm{F}_\theta \left( \bm{s}_t^j, \bm{s}^{\mathcal{O}|j}, \bm{s}^{Adj(j)} \right), t \right),
    \label{eq:calc_flow}
\end{equation}
where $\bm{s}_t^j=(1-t)\bm{s}^j+t\bm{\varepsilon}$ with $\bm{\varepsilon} \in \mathcal{N}(0, \bm{I})$.

\subsubsection{Initialization, training, and sampling.}
\label{sec:init_train_samp}

Similar to many diffusion fine-tuning strategies~\cite{mou2023t2i,zhang2023adding}, our model is initialized to behave identically to the original TRELLIS before fine-tuning, and gradually incorporates the new conditioning as training progresses.
Consequently, we use the pre-trained parameters of the original model for the flow Transformer.
For the proposed MLP layer($\bm{F}_\theta$), the weight matrix is initialized such that its diagonal elements are set to 1, while all other elements and the bias are set to 0.
This ensures that, before training, the output of $\bm{F}_\theta$ is identical to the noise regardless of the condition values.
Then, the flow matching loss~\cite{lipman2023flow} is applied to fine-tune our model.
The loss function to fine-tune the detail enhancer can be written as:
\begin{equation}
    \mathcal{L}_\theta = \mathbb{E}_{\bm{s}^j, t} \lVert \bm{v}_\theta\left(\bm{s}^j, t\right) - (\bm{\varepsilon} - \bm{s}^j) \rVert_2^2.
    \label{eq:calc_loss}
\end{equation}

We note that only the parameters of the MLP layer ($\bm{F}_\theta$) are updated during fine-tuning.

\input{fig/qual_freeform}
\input{fig/qual_syncity}

After fine-tuning, we consecutively apply the model to sample the latent from the noise.
We auto-regressively estimate the structured latent of small cubes from index 0 to 7.
When estimating the latent of cube 0 ($\bm{s}^0$), only the large cube structured latent is used since there is no information on adjacent cubes.
When generating latent for subsequent cubes, the latent of adjacent cubes that have already been generated are also utilized as conditioning inputs.
The eight estimated structured latents are merged and compose a latent that can represent a scene with enhanced details.

It is noteworthy to mention that we do not use classifier-free guidance (CFG)~\cite{ho2021classifierfree} when fine-tuning or sampling the model.
CFG encourages the model to generate same output for both conditional and unconditional settings, and use the difference to guide the denoising direction during sampling; however, in our case, the existence of conditions leads to substantial difference in reconstruction performance, thereby reducing the effectiveness of the CFG strategy.



\subsection{Fine-tuning SLAT Decoder}
\label{sec:method4}
Since the original TRELLIS decoder was trained exclusively on data representing complete objects, its performance degrades when tasked with representing cubes extracted from partial scenes.
To overcome the distribution discrepancy, we fine-tuned the latent decoder ($\mathcal{D}_{L}$) using the small cubes used for training of the detail enhancer.
Specifically, we extracted structured latents from the 3D meshes of these small cubes using the pre-trained encoder. Then we fine-tuned the decoder by minimizing the difference between the reconstructed 3D representation and the original mesh.
The loss function follows the original TRELLIS formulation.
Through decoder fine-tuning, we can obtain higher-quality 3D representations from the structured latents estimated by the detail enhancer.



%% file: fig/method.tex
\begin{figure}
    \includegraphics[width=1.0\textwidth]{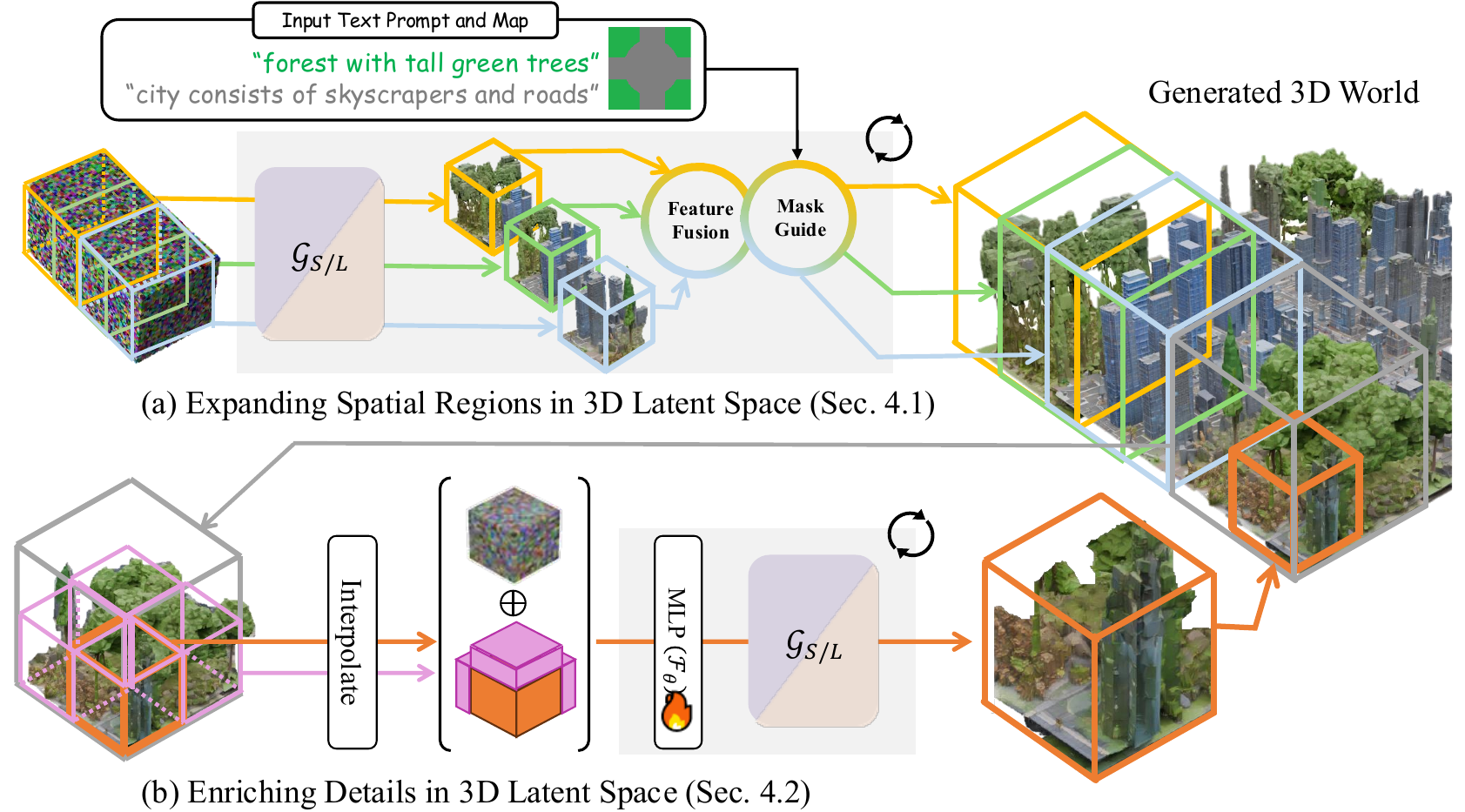}
    \caption{
        Visualization of the overall generation pipeline.
        First, we estimate the structured latent for a large world conditioned by a segmentation map and a text prompt for each segment using latent fusion (top).
        Then, we further upscale the resolution of the generated scene using the detail enhancer.
        Detail enhancer includes an MLP layer to incorporate condition latent and noise, as well as a flow Transformer (bottom).
    }
    \label{fig:method}
\end{figure}

%% file: fig/qual_freeform.tex
\begin{figure*}[t]
    \newcommand{\w}{0.175\linewidth}
    \newcommand{\ww}{0.35\linewidth}
    \newcommand{\www}{0.28\linewidth}
    \newcommand{\hh}{0.175\linewidth}
    \centering

    \subfloat{\includegraphics[width=\ww, height=\hh]{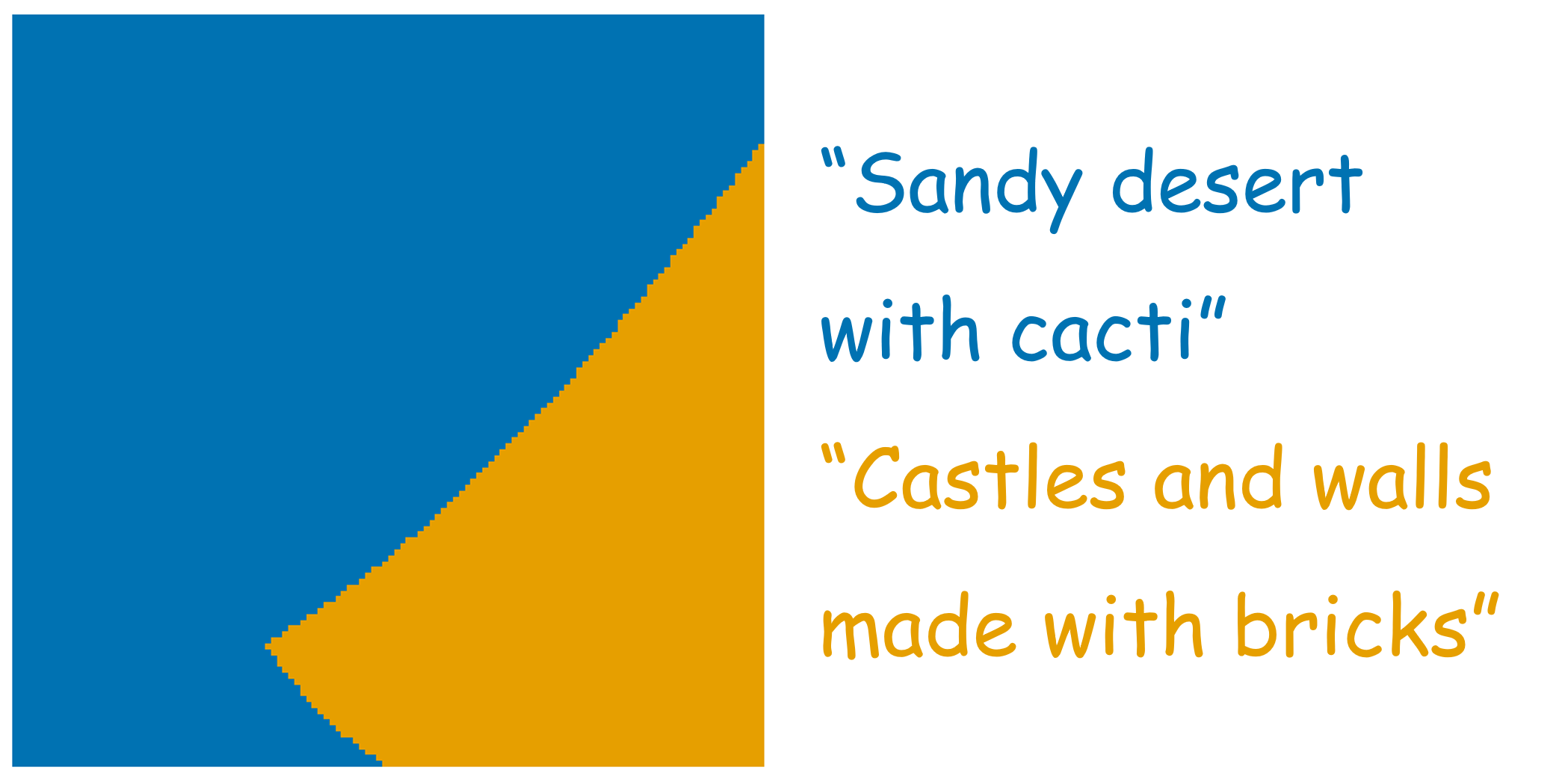}}
    \hfill
    \subfloat{\includegraphics[width=\w, height=\hh]{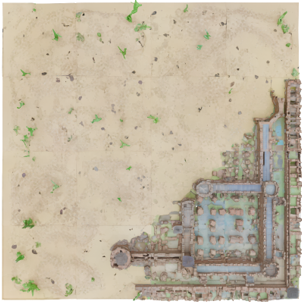}}
    \hfill
    \subfloat{\includegraphics[width=\www, height=\hh]{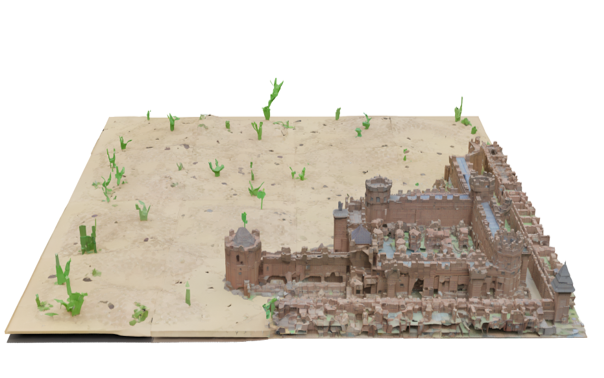}}
    \hfill
    \subfloat{\includegraphics[width=\w, height=\hh]{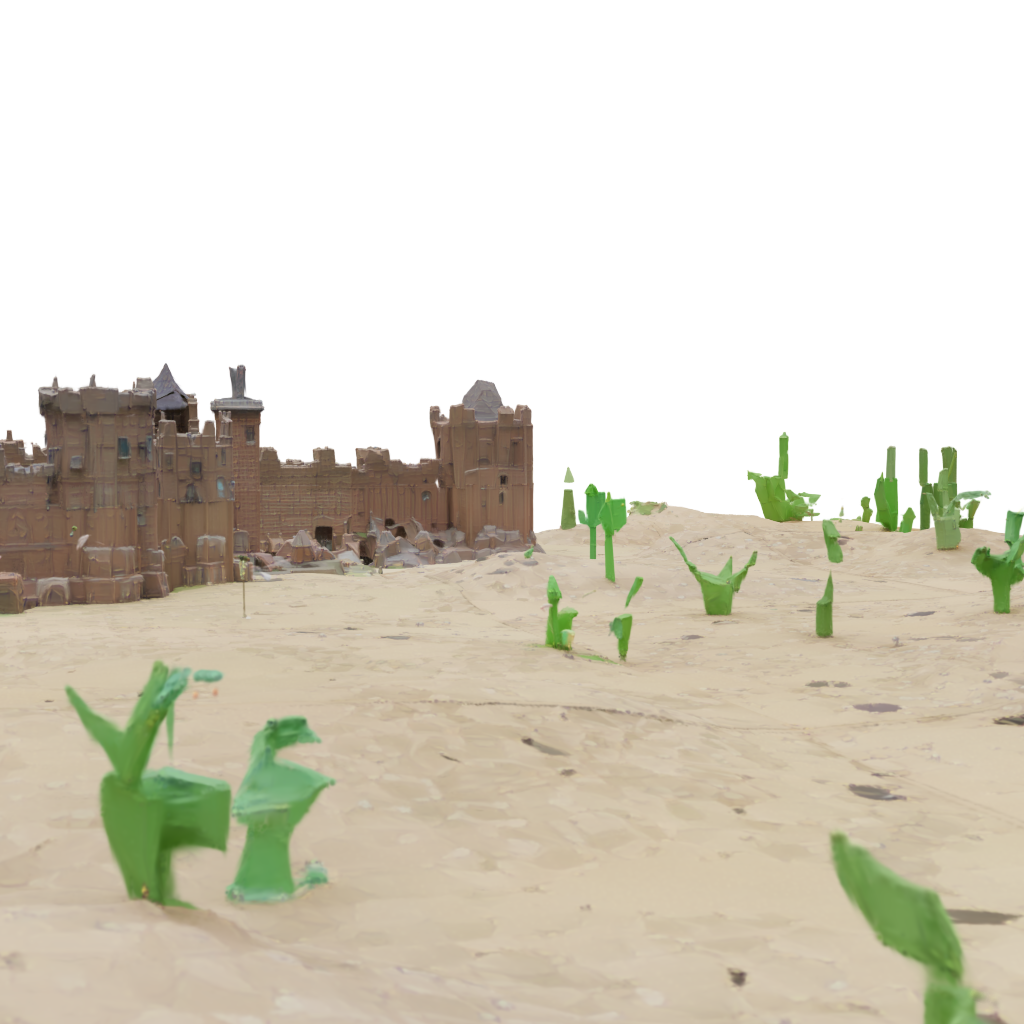}}

    \vspace{0.1cm}
    
    \addtocounter{subfigure}{-4}
    \subfloat{\includegraphics[width=\ww, height=\hh]{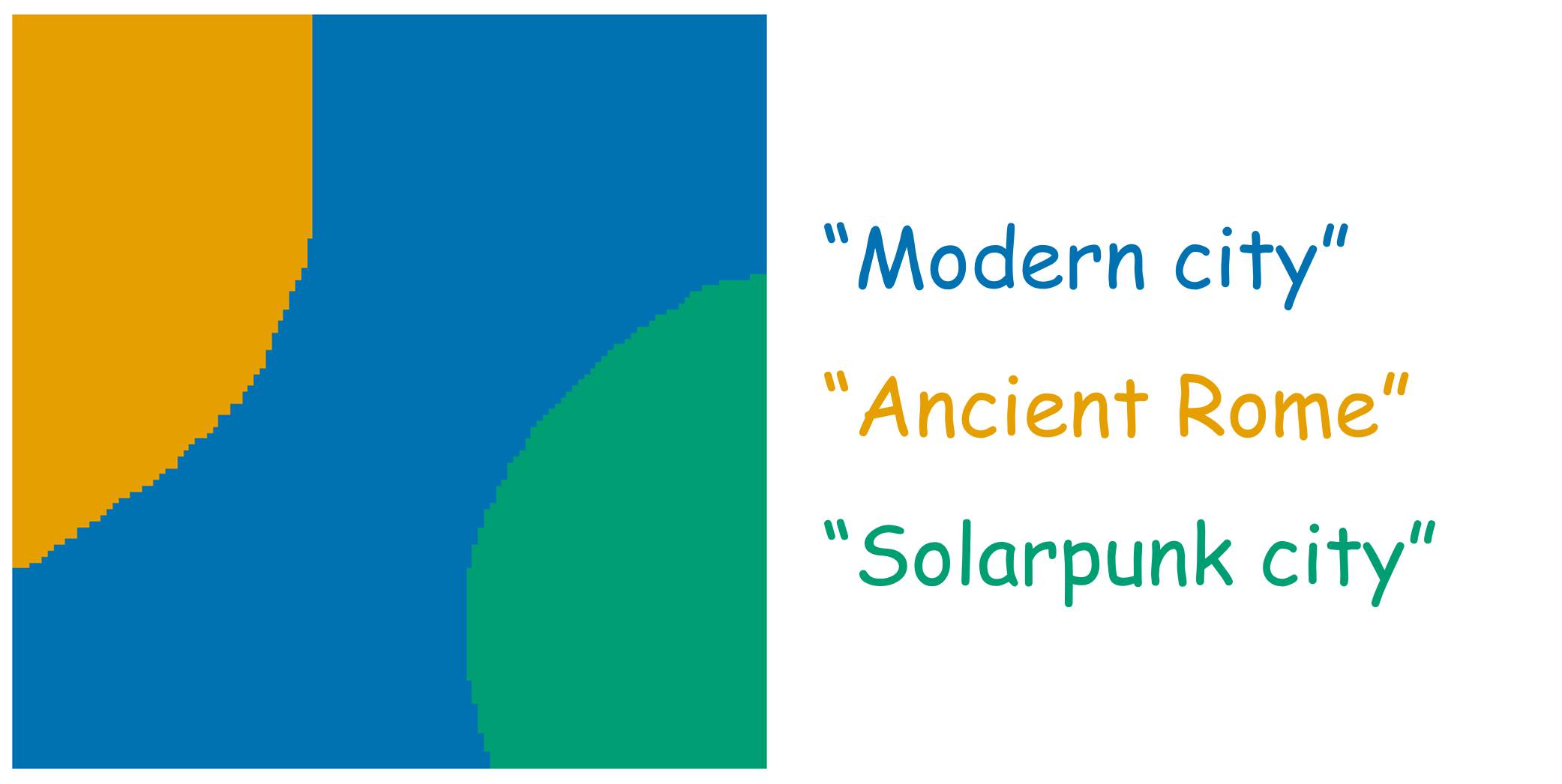}}
    \hfill
    \subfloat{\includegraphics[width=\w, height=\hh]{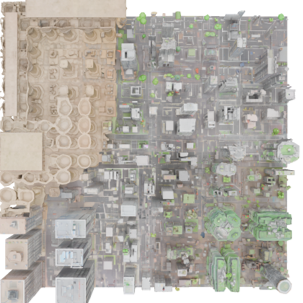}}
    \hfill
    \subfloat{\includegraphics[width=\www, height=\hh]{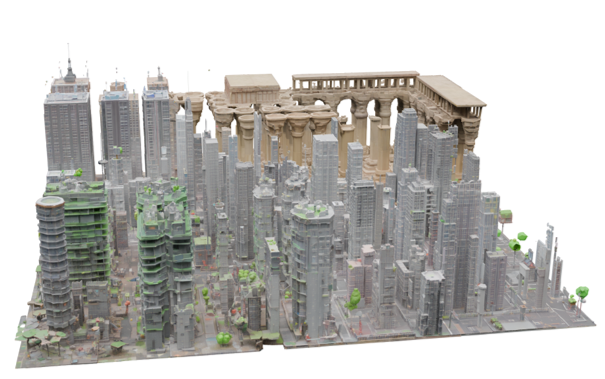}}
    \hfill
    \subfloat{\includegraphics[width=\w, height=\hh]{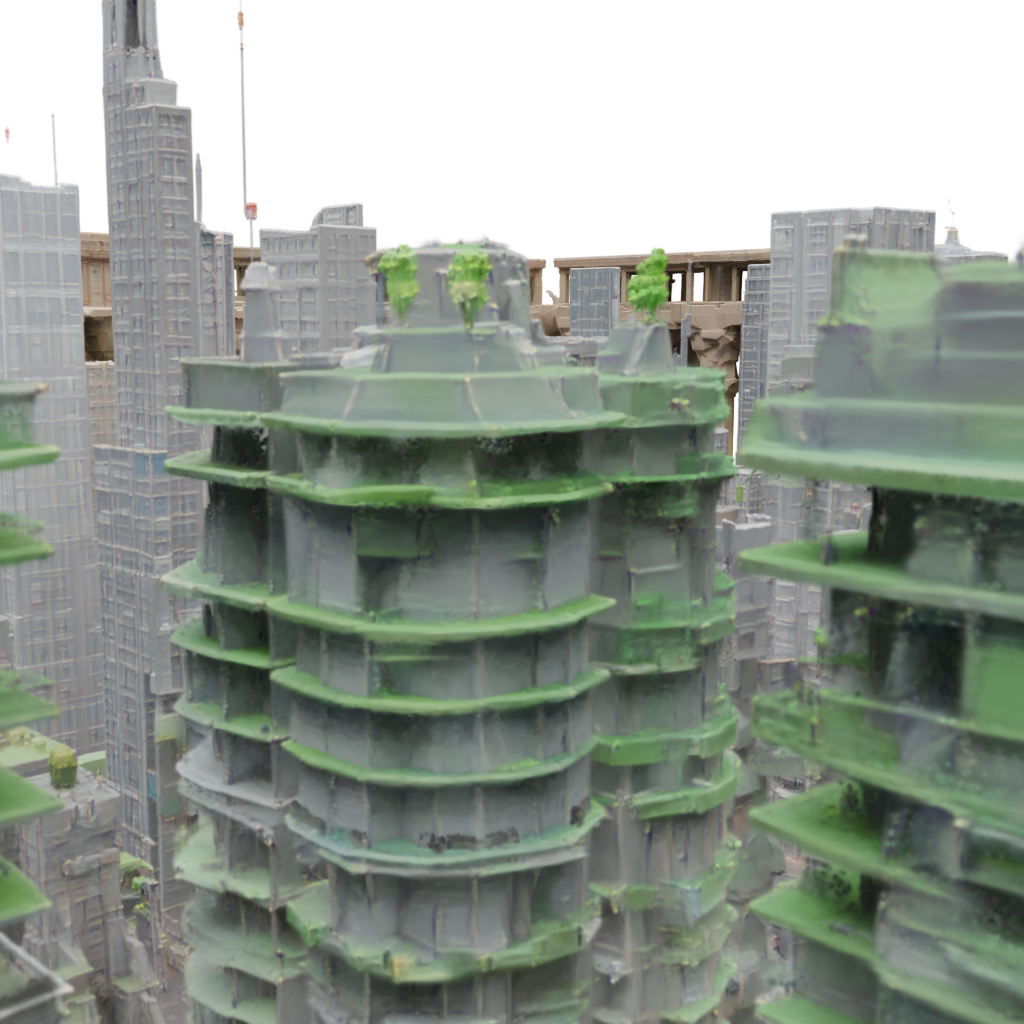}}

    \vspace{0.1cm}
    
    \addtocounter{subfigure}{-4}
    \subfloat[Segment map]{\includegraphics[width=\ww]{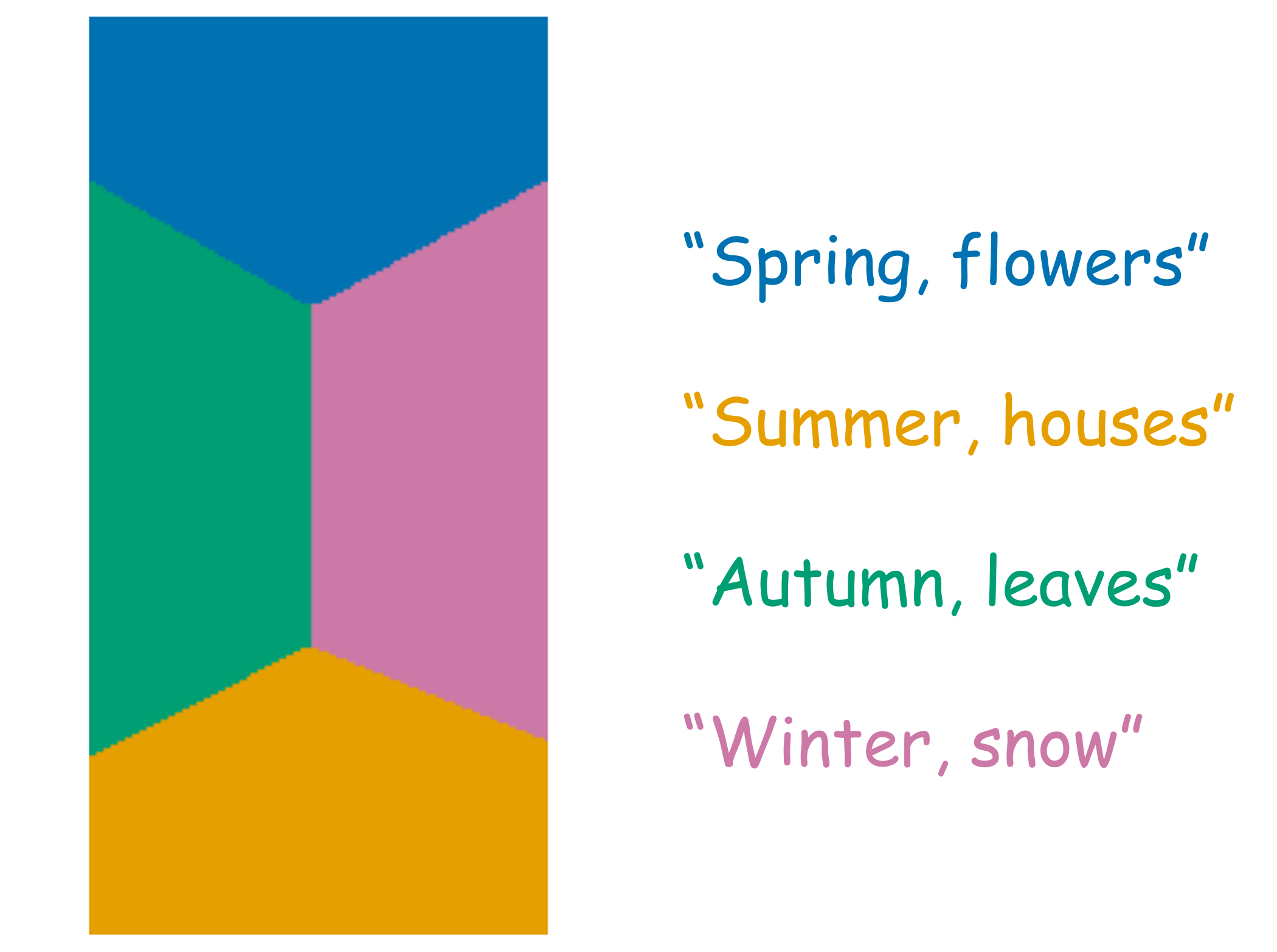}\label{fig:qual_freeform_a}} 
    \hfill
    \subfloat[Top-view map]{\includegraphics[width=\w]{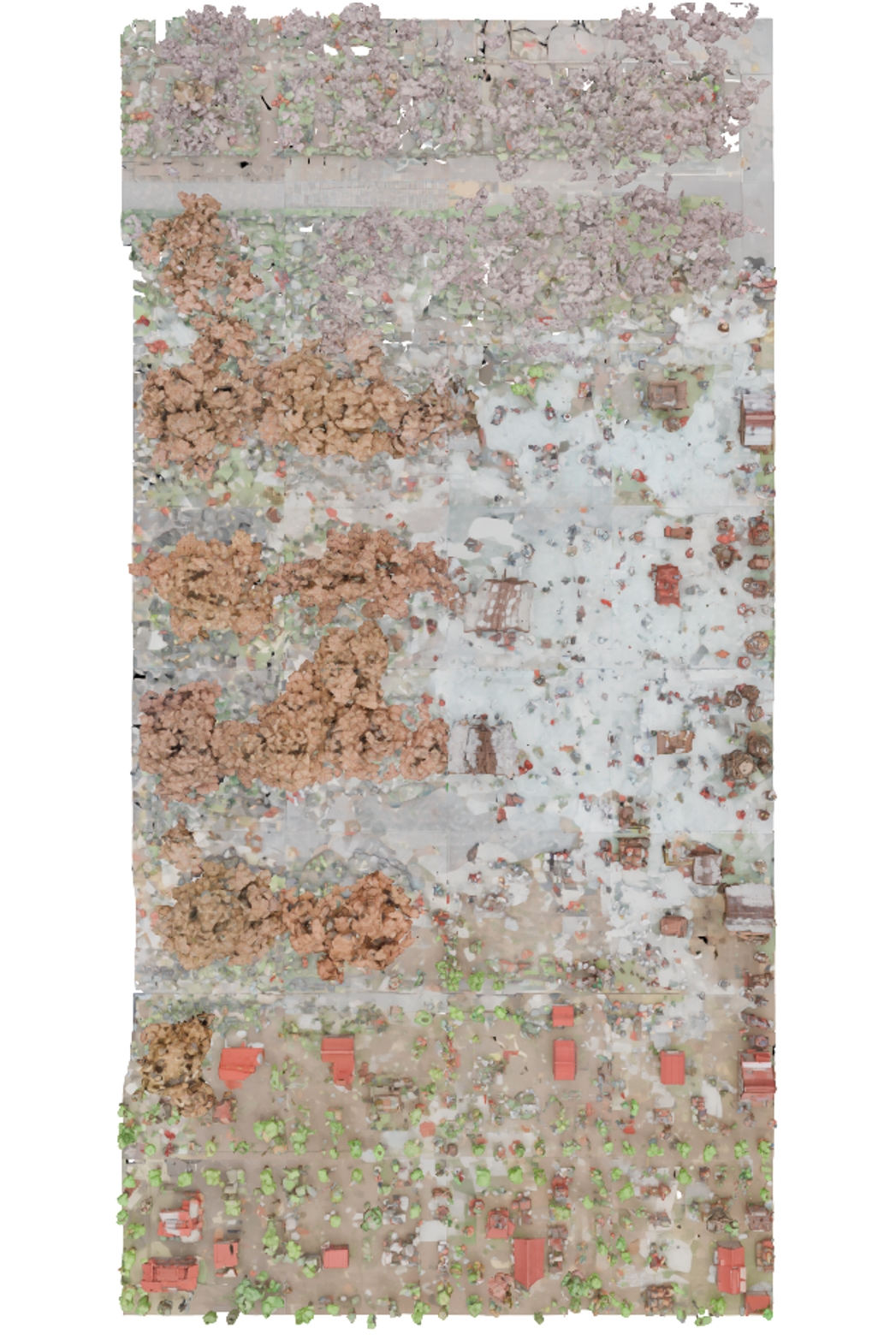}\label{fig:qual_freeform_b}} 
    \hfill
    \subfloat[Generated world]{\includegraphics[width=\www]{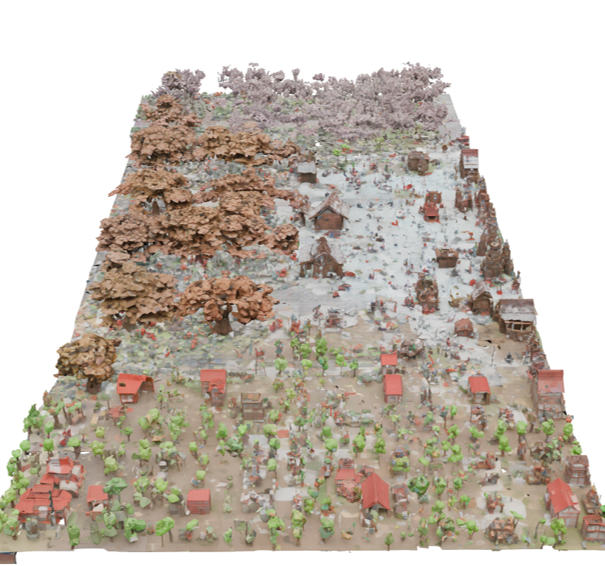}\label{fig:qual_freeform_c}}
    \hfill
    \subfloat[Roaming view]{\includegraphics[width=\w]{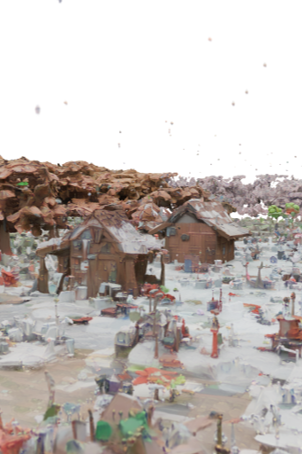}\label{fig:qual_freeform_d}}
    
    \caption{
        Qualitative results conditioned by arbitrary shaped segment maps with user-defined text prompts.
        Our model creates the 3D world corresponds to the input map regardless of its size and shape.
        We note that \textbf{Syncity cannot generate the scene from such a complicated segment map}.
    }
    \label{fig:qual_freeform}
\end{figure*}

%% file: fig/qual_syncity.tex
\begin{figure*}[t]
    \centering
    \newlength{\LeftTallH}  
    \newlength{\RowH}       
    \newlength{\RowGap}     
    
    \setlength{\RowH}{0.23\textheight}   
    \setlength{\RowGap}{0.012\textheight} 
    \setlength{\LeftTallH}{\dimexpr 2\RowH + \RowGap\relax}

  \begin{minipage}[b]{0.34\linewidth}
    \centering
    \begin{subfigure}[t]{\linewidth}
      \centering
      \includegraphics[width=\linewidth,height=\LeftTallH,keepaspectratio]{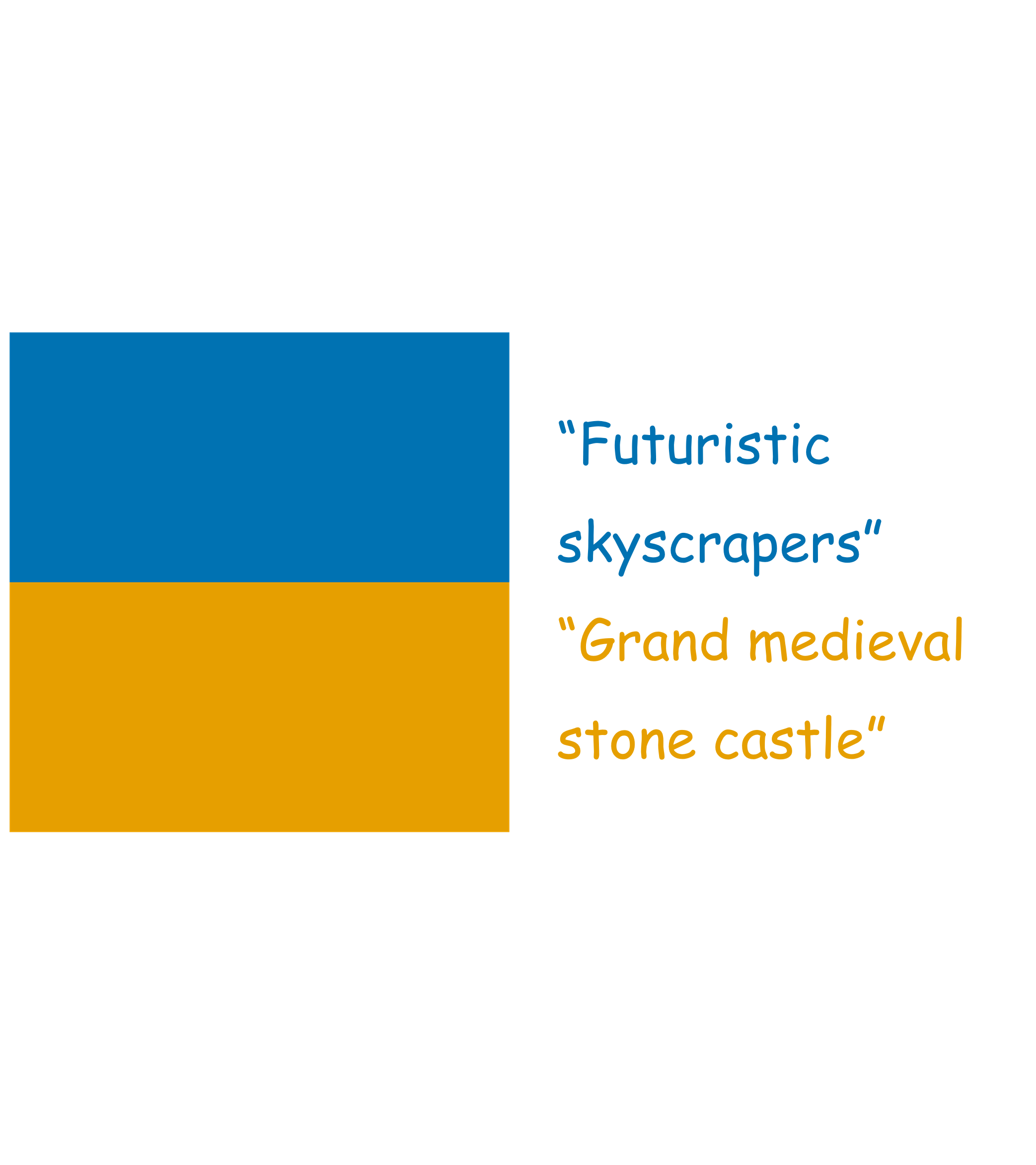}
      \caption{Segment map}
      \label{fig:left-tall}
    \end{subfigure}
  \end{minipage}
  \hfill
  \begin{minipage}[b]{0.65\linewidth}
    \begin{minipage}[b]{0.25\linewidth}
      \centering
      \begin{subfigure}[t]{\linewidth}
        \centering
        \includegraphics[width=\linewidth,height=\RowH,keepaspectratio]{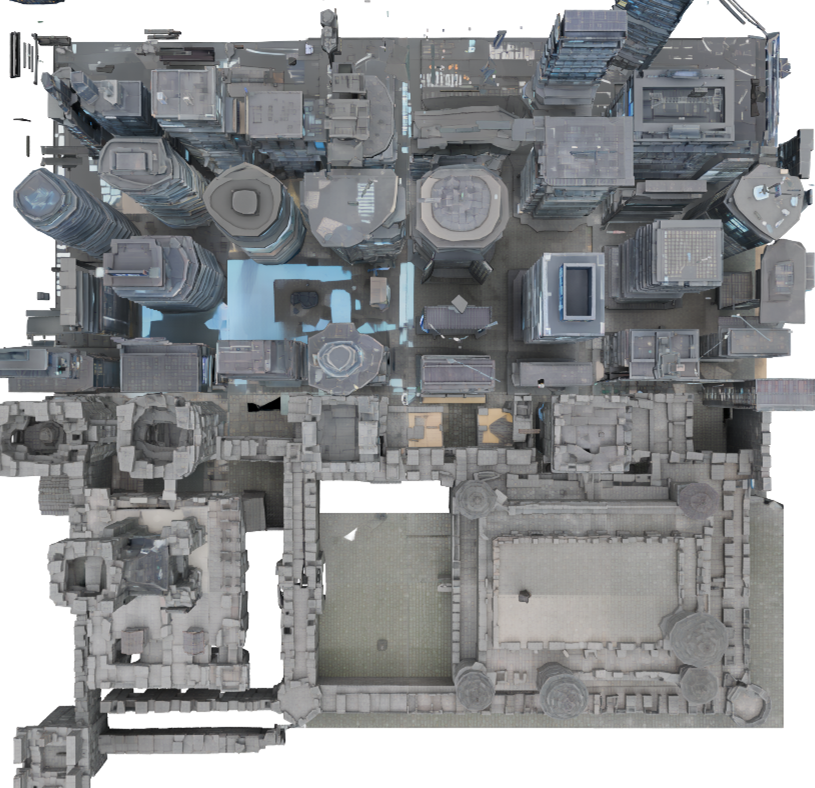}
        \caption{Ours - map}
        \label{fig:top-left}
      \end{subfigure}
    \end{minipage}%
    \hfill
    \begin{minipage}[b]{0.41\linewidth}
      \centering
      \begin{subfigure}[t]{\linewidth}
        \centering
        \includegraphics[width=\linewidth,height=\RowH,keepaspectratio]{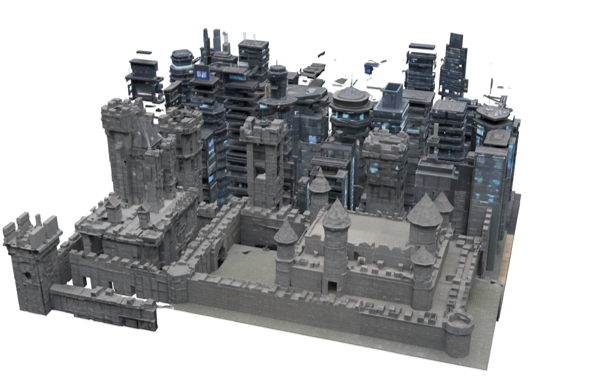}
        \caption{Ours - generated world}
        \label{fig:top-right}
      \end{subfigure}
    \end{minipage}
    \hfill
    \begin{minipage}[b]{0.25\linewidth}
      \centering
      \begin{subfigure}[t]{\linewidth}
        \centering
        \includegraphics[width=\linewidth,height=\RowH,keepaspectratio]{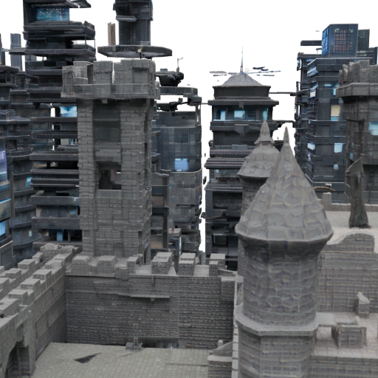}
        \caption{Ours - roaming view}
        \label{fig:top-right}
      \end{subfigure}
    \end{minipage}
    
    \vspace{\RowGap}

    \begin{minipage}[b]{0.25\linewidth}
      \centering
      \begin{subfigure}[t]{\linewidth}
        \centering
        \includegraphics[width=\linewidth,height=\RowH,keepaspectratio]{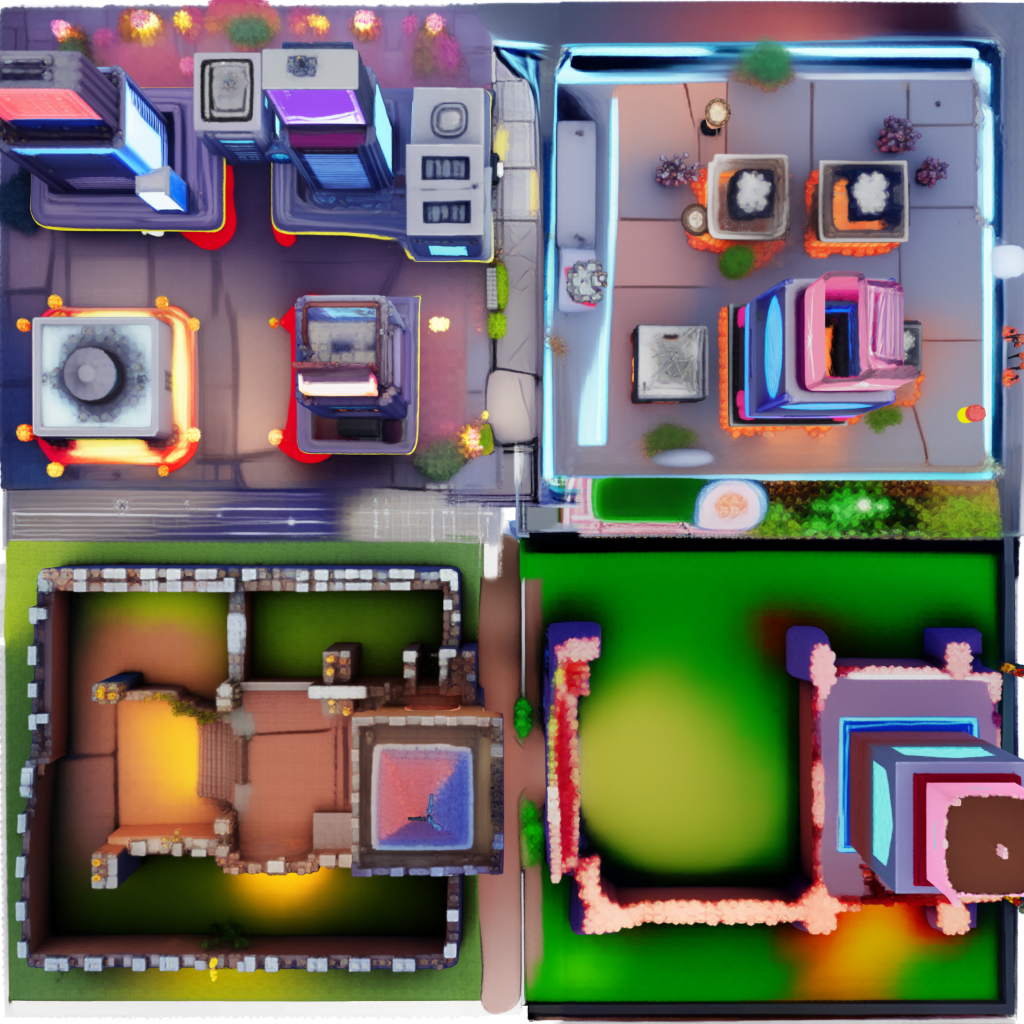}
        \caption{SynCity - map}
        \label{fig:bottom-left}
      \end{subfigure}
    \end{minipage}%
    \hfill
    \begin{minipage}[b]{0.41\linewidth}
      \centering
      \begin{subfigure}[t]{\linewidth}
        \centering
        \includegraphics[width=\linewidth,height=\RowH,keepaspectratio]{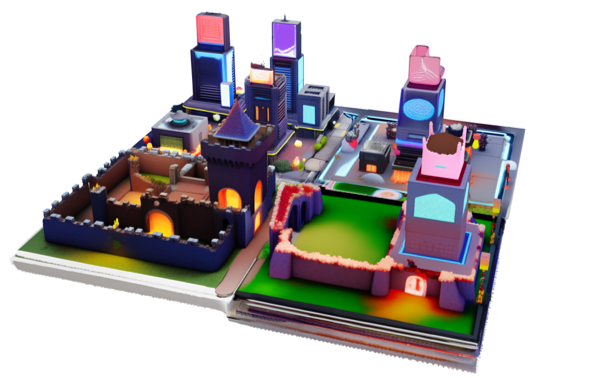}
        \caption{SynCity - generated world}
        \label{fig:bottom-right}
      \end{subfigure}
    \end{minipage}
    \hfill
    \begin{minipage}[b]{0.25\linewidth}
      \centering
      \begin{subfigure}[t]{\linewidth}
        \centering
        \includegraphics[width=\linewidth,height=\RowH,keepaspectratio]{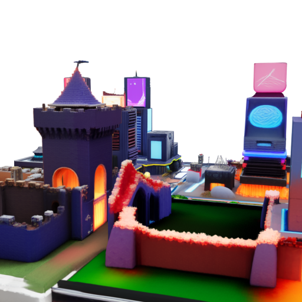}
        \caption{SynCity - roaming view}
        \label{fig:bottom-right}
      \end{subfigure}
    \end{minipage}
    
  \end{minipage}
    
    \caption{
        Qualitative comparison with SynCity in a grid-type segmentation map.
        Syncity only supports generating assets on square-type tiles, and it lacks contextual continuity between adjacent tiles.
        Our model fills the entire world with appropriate objects in a seamless manner.
    }
    \label{fig:qual_syncity}
\end{figure*}

%% file: sec/4_experiment.tex
\section{Experiments} \label{sec:experiment}

\subsubsection{Dataset curation.}
Using the labels of NuiScene43~\cite{lee2025nuiscene}, we choose a set of filtered 43 high-quality scene meshes from Objaverse~\cite{deitke2023objaverse} to train the detail enhancers.
We exclude eight scenes without textural information and extract pairs from the remaining 35 scenes.
We randomly cropped 500 cubes with various sizes ($\in\{64, 128, 192, 256\}$) for each scene.
Here, we checked that each cube includes more than 10,000 vertices.
After a total of 17,500 cubes have been extracted, we treat each cube as a large cube and split it into eight identical small cubes.
For a large cube and the corresponding small cubes, we extract sparse structure latent and structured latent tensors using the pre-trained encoder models of TRELLIS.
We use 16,000 pairs to train the enhancer and 1,500 pairs for validation.

\subsection{Comparison on World Generation}

Map2World takes user-provided text prompts and a segmentation map comprising multiple segments labeled by each prompt as input, and generates a large-scale world that satisfies these conditions.
We note that our model is not restricted to the size of the voxel grid of the original asset generator and can create the world with any user-defined dimensions.
Also, the region of each segment can be formed in an arbitrary shape.
\cref{fig:qual_freeform} displays the rendered samples of the world created by Map2World, given the segment map with free-form segments.
The three examples shown in the figure differ not only in their input prompts but also in the number of segments, the segment shapes, and the overall dimensions of the target world.
SynCity cannot generate an appropriate 3D world under these conditions.
In contrast, our model successfully produces a 3D world that aligns well with both the segment map shapes and the text prompts in all cases.
\cref{fig:qual_freeform_b} displays a top-down view of the generated world, where each region closely matches the input segment map.
Furthermore, as illustrated in \cref{fig:qual_freeform_c,fig:qual_freeform_d}, the structures of adjacent regions are seamlessly connected, ensuring smooth transitions between segments.

\cref{fig:qual_syncity} shows rendered views of the generated world where the map consists of grid-shaped segments.
We compare the generated world by Map2World with SynCity~\cite{engstler2025syncity} that also uses TRELLIS for scene generation.
Our model surpasses SynCity in two key aspects.
First, our method can generate a large connected structure, while SynCity cannot create large objects that occupy multiple tiles with the same labels.
Moreover, for SynCity, the coherence among generated objects is weak, and gaps between tiles are often left unfilled, giving the impression of multiple disconnected assets placed together.
Second, our model produces a relatively denser and more complex world.
Even if SynCity shows better inter-object coherence, the world looks like a grid-like arrangement of assets with numerous empty spaces, which is not considered a realistic example.
In contrast, Map2World generates large and structurally complex assets within each segment and naturally connects the contents between adjacent segments.
As shown in the rendering example on the right columns of \cref{fig:qual_syncity}, our results appear significantly closer to a real-world environment.

For a quantitative comparison, we generated text captions for 35 meshes from NuiScene43 and synthesized scenes from each caption using both SynCity and our proposed method. The rendered images were then evaluated across four criteria — sharpness, world completeness, coherence, and realism — using GPTscore~\cite{fu2024gptscore}. When averaged across all images, our model achieved a score of 7.93/10, outperforming SynCity's 7.48/10, indicating that our approach generates more complete and high-quality 3D worlds.


Moreover, we evaluate generated environments using a composite metric, \emph{World Quality (WQ)}, defined as
\begin{equation}
WQ = 0.15S + 0.45W + 0.25C + 0.15R .
\end{equation}
Here, $S$ denotes \emph{sharpness}, measuring the visual fidelity and clarity of geometric edges in the rendered scene.
$W$ represents \emph{world completeness and complexity}, evaluating the scale of the environment, the richness of spatial structures, and the diversity of scene elements.
$C$ denotes \emph{coherence and consistency}, assessing whether the generated layout forms a structurally consistent world, such as aligned roads, plausible spatial relationships, and the absence of geometric conflicts.
$R$ represents \emph{realism}, measuring the overall plausibility of the scene, including lighting, materials, and geometric plausibility.
All metrics are measured using the GPT 5.3 model.

The proposed \emph{WQ} metric emphasizes the structural quality of the generated world rather than the fidelity of individual objects.
By assigning the highest weight to world completeness, the metric prioritizes large-scale environmental structure and layout complexity, while coherence and realism ensure that increased scale does not come at the cost of structural inconsistency or visual artifacts.
This design allows $WQ$ to better capture the quality of generated environments as coherent worlds, making it particularly suitable for evaluating world-scale generative models.

\input{supp/table/gptscore}
\cref{tab:supp_gptscore} shows the samples of generated results with the same text prompt and calculated \emph{WQ} metric using the same input prompt with different generation models.
We selected GaussianCube~\cite{zhang2024gaussiancube} and SynCity~\cite{engstler2025syncity} as the comparison frameworks.
GaussianCube produces environments with noticeably lower visual quality and incomplete scene structures, resulting in low scores across most evaluation criteria.
While SynCity benefits from the strong image-to-3D expressive capability of TRELLIS, thereby achieving high sharpness, the generated assets are often connected in an unnatural manner.
These inconsistencies reduce the structural completeness of the generated environments and lead to a relatively lower world completeness score.
In contrast, Map2World achieves consistently high scores across all evaluation criteria, producing environments that are not only visually clear but also structurally coherent and complete at the world level.

\subsection{Ablation Studies}

\input{fig/ablation_initnoise}
\input{fig/ablation_initnoise_graph}
\input{fig/ablation_model}

\subsubsection{Spectral parameterization for stable initial latent optimization.}
\label{sec:ablation:noise}

In the iterative optimization of initial noise for scale control illustrated in \cref{fig:ablation_init}, the repeated generation increases both generation time and computational cost.
Therefore, we introduce the spectral domain parameterization of the sparse structure $S$ as described in \cref{sec:optinit}, to stabilize the optimization trajectory, enable the use of a large learning rate, and enrich the target scale within a few optimization steps.
We present the IoU and Dice plots measuring how well the target geometry constraint is satisfied in \cref{fig:ablation_init_graph}.
In our setting, the blue curve shows that with a high learning rate of 9.0, the IoU and Dice scores reach approximately 0.9 on average within five optimization steps. In contrast, the orange curve, which directly optimizes the sparse structure without spectral-domain parameterization under the same setting, exhibits unstable behavior with large loss spikes and fails to converge. Reducing the learning rate to 1.0 (green curve) avoids divergence but requires significantly more optimization steps, increasing the computational burden.


\subsubsection{Design choices for the detail enhancer.}
%

We conduct ablation studies on the design of the detail enhancer and decoder fine-tuning to justify our design choice.
\cref{fig:ablation_model} shows rendered images from generated scenes by changing the options while keeping the condition, seed, and camera parameters the same.
\cref{fig:ablation_model_e} is the rendered sample without using the detail enhancer; the generated structured latent with latent fusion is decoded and rendered.

\input{table/recon_expand}

First, we apply different network architectures to compare the performance.
Specifically, we adapt the architecture from \textbf{IP-Adapter}~\cite{ye2023ip-adapter,mou2023t2i} to fine-tune $\mathcal{G}_{S/L}$.
The IP-adapter is also designed for fine-tuning the model to the new condition in a parameter-efficient manner.
However, as shown in \cref{fig:ablation_model_b}, IP-Adapter fails to generate the appropriate structure and shows disconnectedness at the boundary.
Also, we use \textbf{classifier-free guidance} during fine-tuning and sampling.
The sampled result is shown in \cref{fig:ablation_model_c}.
In our setting, the difference in generation quality between the conditional and unconditional models would be substantial.
When denoising with CFG, the difference between the denoised features in the conditional and unconditional settings becomes excessively large, causing severe geometric distortions and overly saturated colors.
Finally, we analyze the effect of \textbf{SLAT decoder fine-tuning} in \cref{fig:ablation_model_d}.
Although the improvement was less pronounced compared to the changes in the detail enhancer, we observed that the fine-tuned decoder produced sharper geometry and more refined textures.

To quantitatively compare the options, we use the test meshes and consider the rendered images from the mesh as ground truth.
For the metrics, we compute PSNR, LPIPS, and FID scores on the rendered images from the test-set generated scenes and record the results in \cref{tab:recon}.
For FID, we measured the scores across three baseline models: Inceptionv3, DINOv2, and CLIP.
We note that our detail enhancer is a generator; thus, PSNR and LPIPS are not ideal metrics for evaluating quality.
Still, our model achieves the best PSNR and LPIPS among the options, indicating the highest fidelity to the input condition.
Our choice also yields the best FID metrics across all three baselines, suggesting that our model generates the most natural and high-quality detailed structures and textures.

%% file: supp/table/gptscore.tex
\begin{table}[t]
\centering
\caption{Evaluation using the proposed World Quality (WQ) metric where 
$WQ = 0.15S + 0.45W + 0.25C + 0.15R$.}
\begin{tabular}{l|ccc}
\toprule
Images / Score & GaussianCube & SynCity & Map2World \\
\midrule
\multirow{2}{*}{Sample Images} & \includegraphics[width=2.5cm]{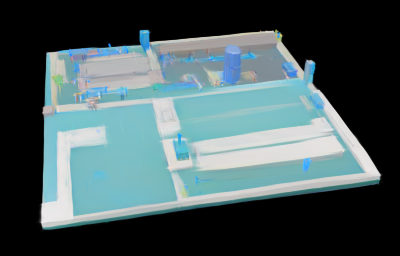} & \includegraphics[width=2.5cm]{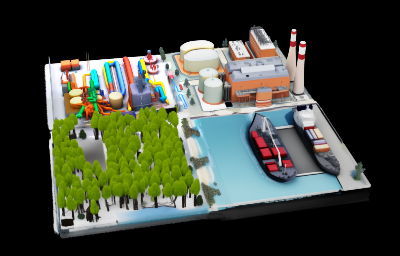} & \includegraphics[width=2.5cm]{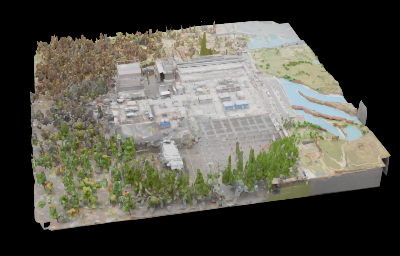} \\
 & \includegraphics[width=2.5cm]{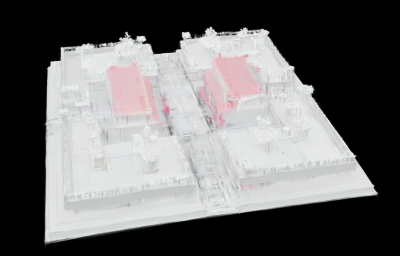} & \includegraphics[width=2.5cm]{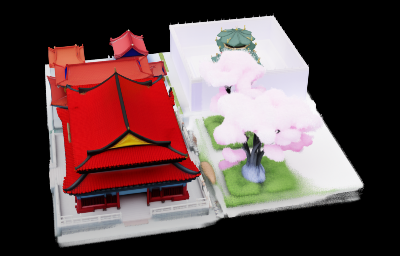} & \includegraphics[width=2.5cm]{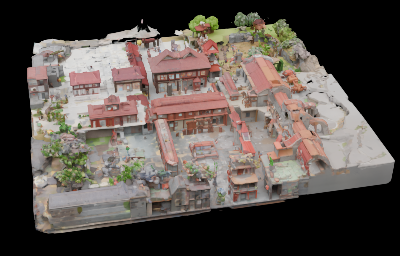} \\
\midrule
$S$ (sharpness) & 6.8 & 8.2 & 8.0 \\
$W$ (world completeness) & 4.5 & 6.8 & 7.8 \\
$C$ (coherence) & 5.0 & 7.6 & 7.9 \\
$R$ (realism) & 5.1 & 7.3 & 7.6 \\
\midrule
$WQ$ & 5.08 & 7.25 & \textbf{7.76} \\
\bottomrule
\end{tabular}
\label{tab:supp_gptscore}
\end{table}

%% file: fig/ablation_initnoise.tex
\begin{figure}[t!]
    \centering
    \includegraphics[width=1.0\linewidth]{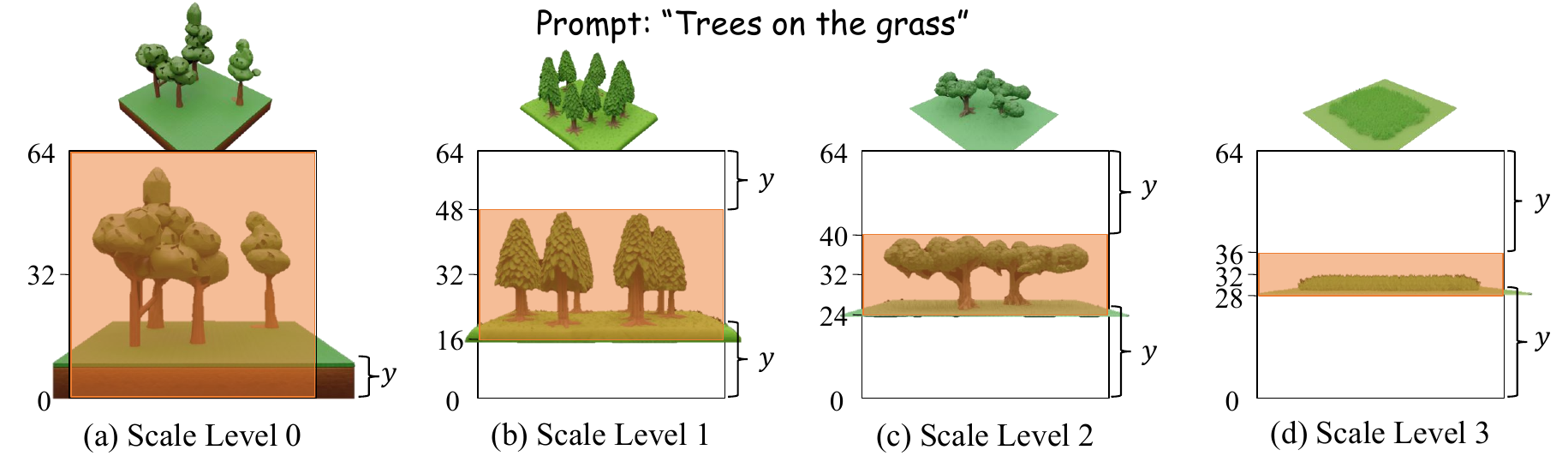}
    \caption{Samples in various scales generated by the same prompt but different seeds.}
    \label{fig:ablation_init}
\end{figure}

%% file: fig/ablation_initnoise_graph.tex
\begin{figure}[t!]
    \vspace{-1mm}
    \centering
    \includegraphics[width=1.0\linewidth]{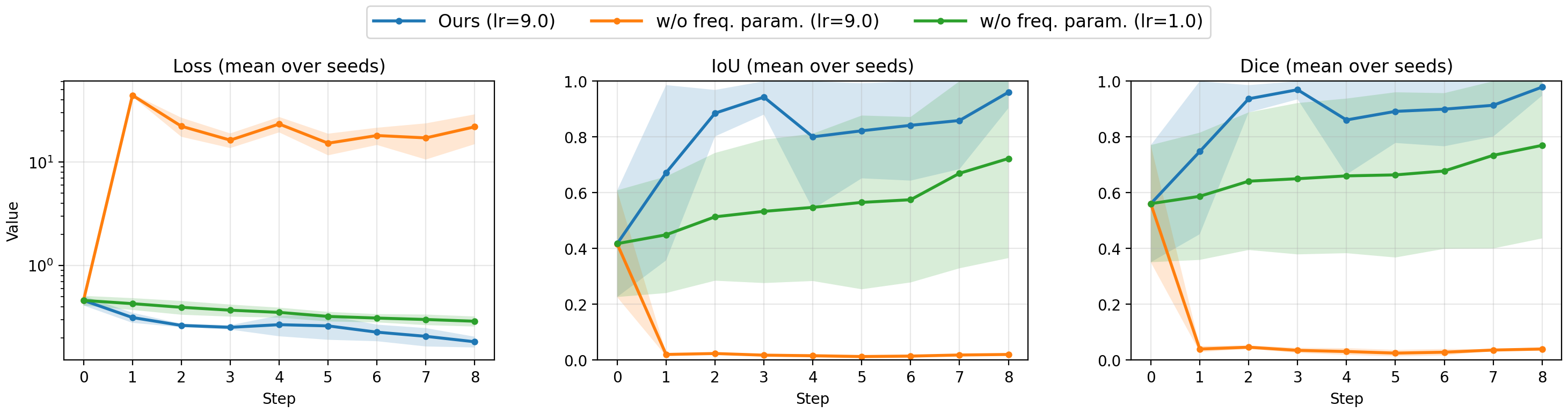}
    \caption{Ablation on spectral-domain parameterization. This parameterization stabilizes the optimization process, enabling rapid convergence to the target within a few steps using a high learning rate.}
    \label{fig:ablation_init_graph}
\end{figure}

%% file: fig/ablation_model.tex
\begin{figure*}[ht]
    \newcommand{\ww}{0.195\linewidth}
    \centering

    \subfloat[Ours]{\includegraphics[width=\ww]{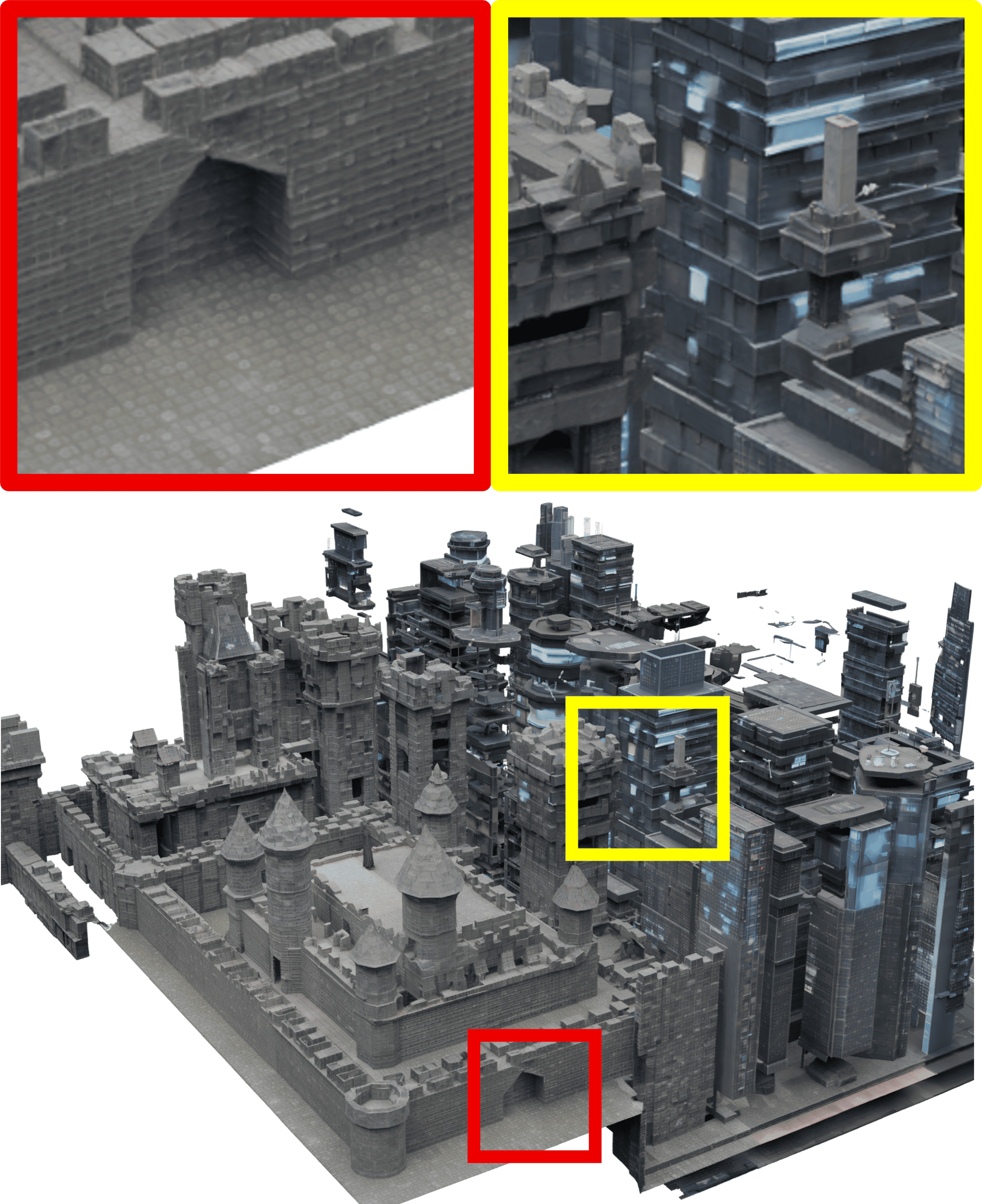}\label{fig:ablation_model_a}}
    \hfill
    \subfloat[IP-Adapter]{\includegraphics[width=\ww]{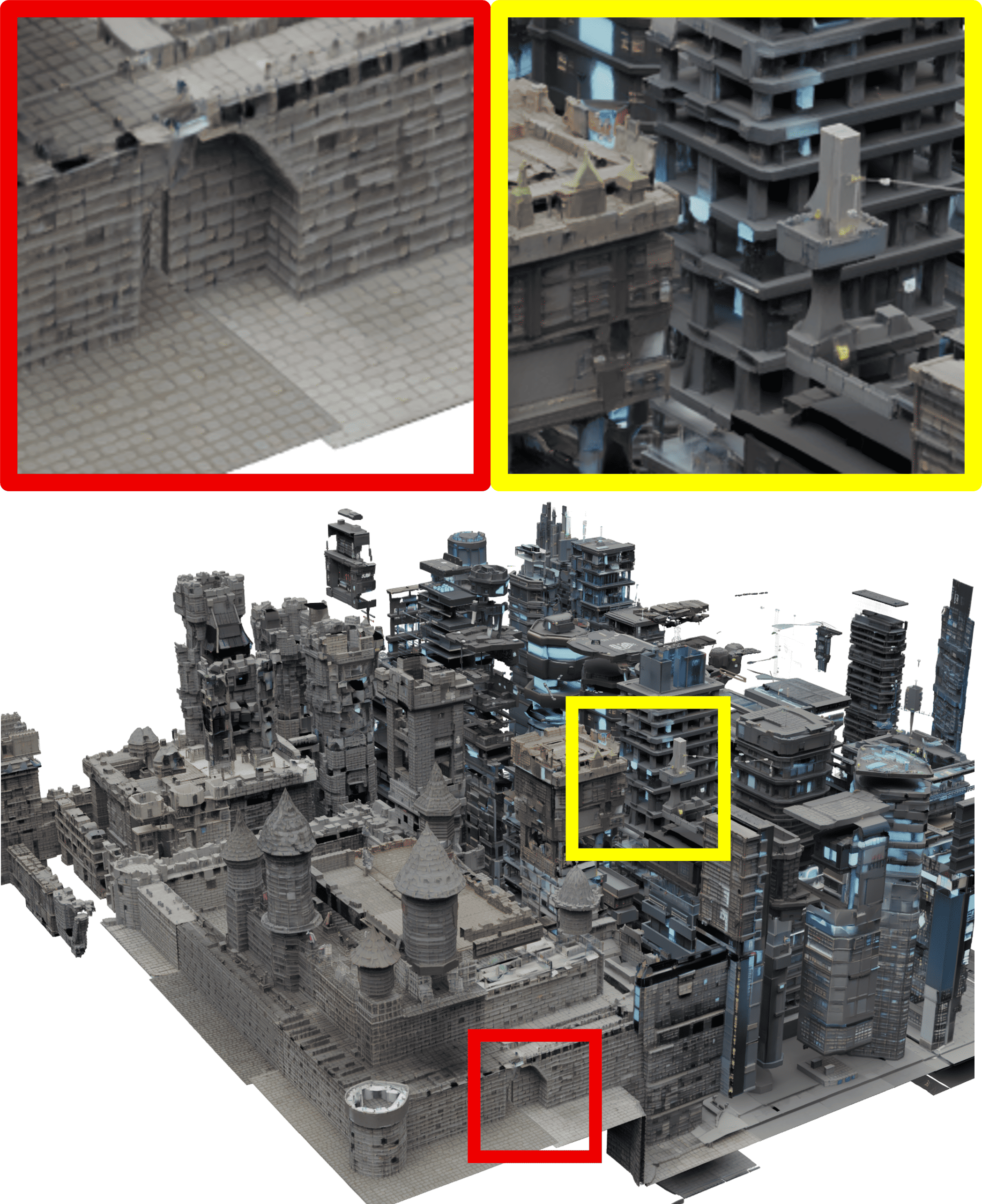}\label{fig:ablation_model_b}} 
    \hfill
    \subfloat[Use CFG]{\includegraphics[width=\ww]{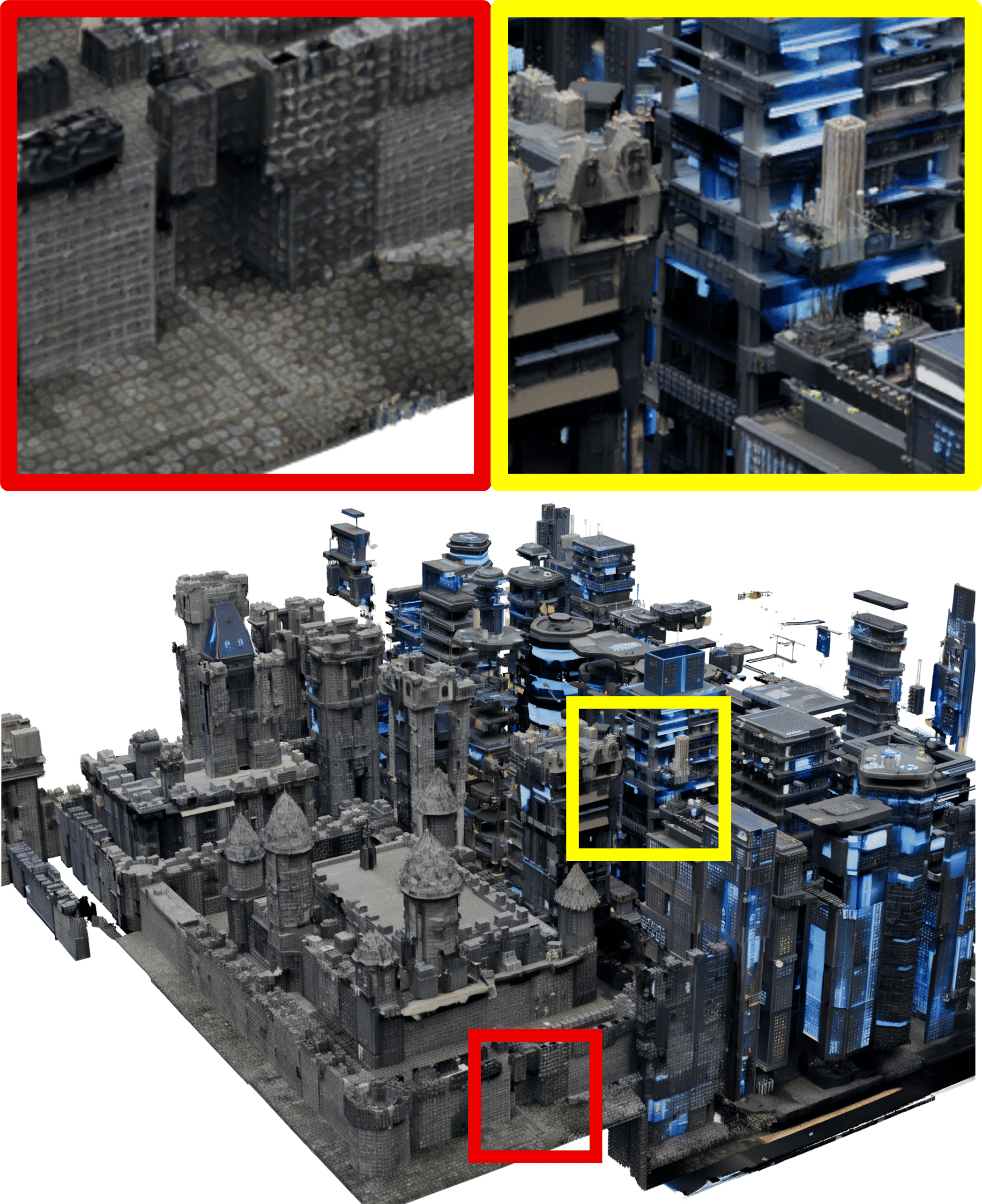}\label{fig:ablation_model_c}}
    \hfill
    \subfloat[No $\mathcal{D}_L$ fine-tuning]{\includegraphics[width=\ww]{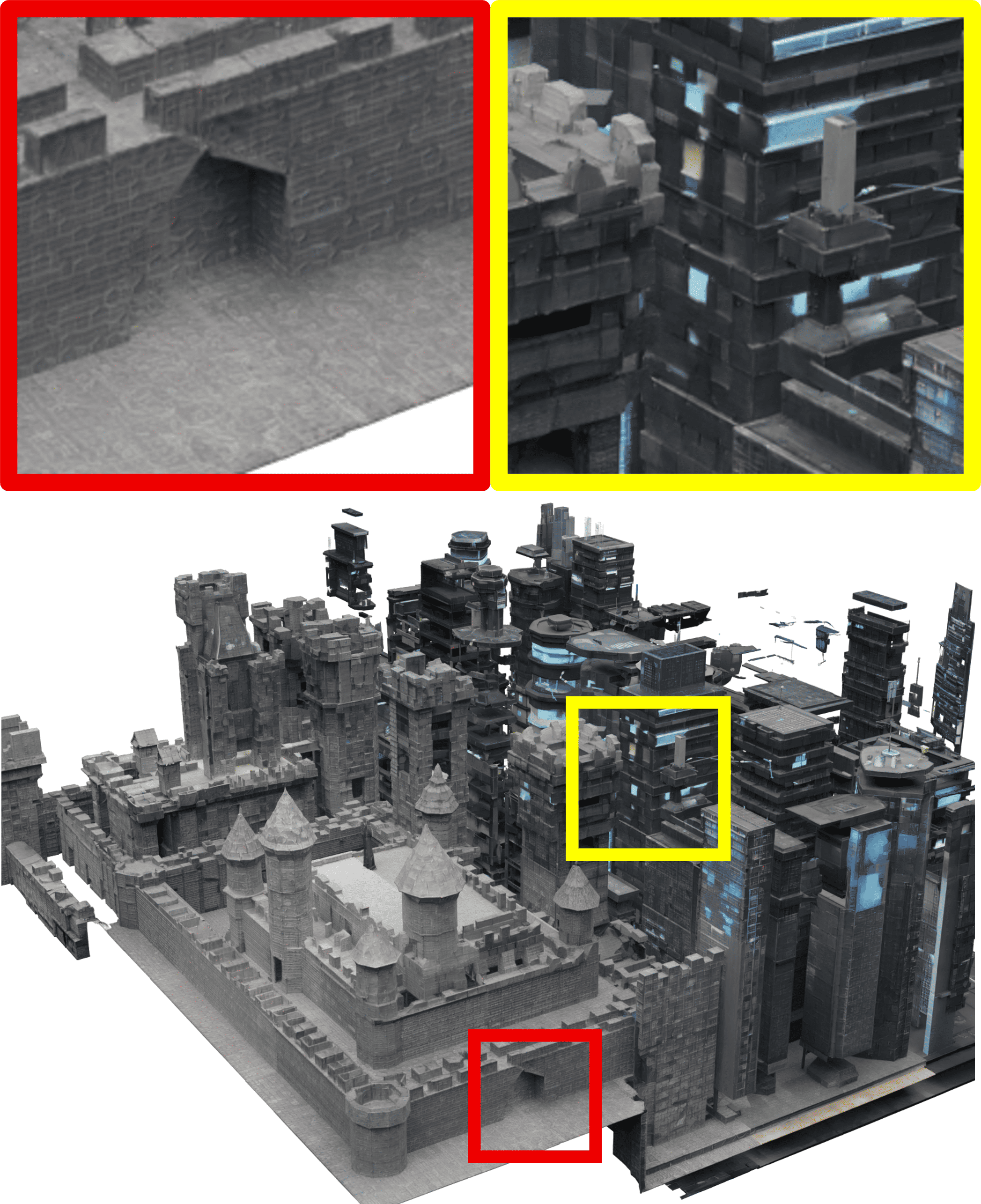}\label{fig:ablation_model_d}}
    \hfill
    \subfloat[No detail enhancing]{\includegraphics[width=\ww]{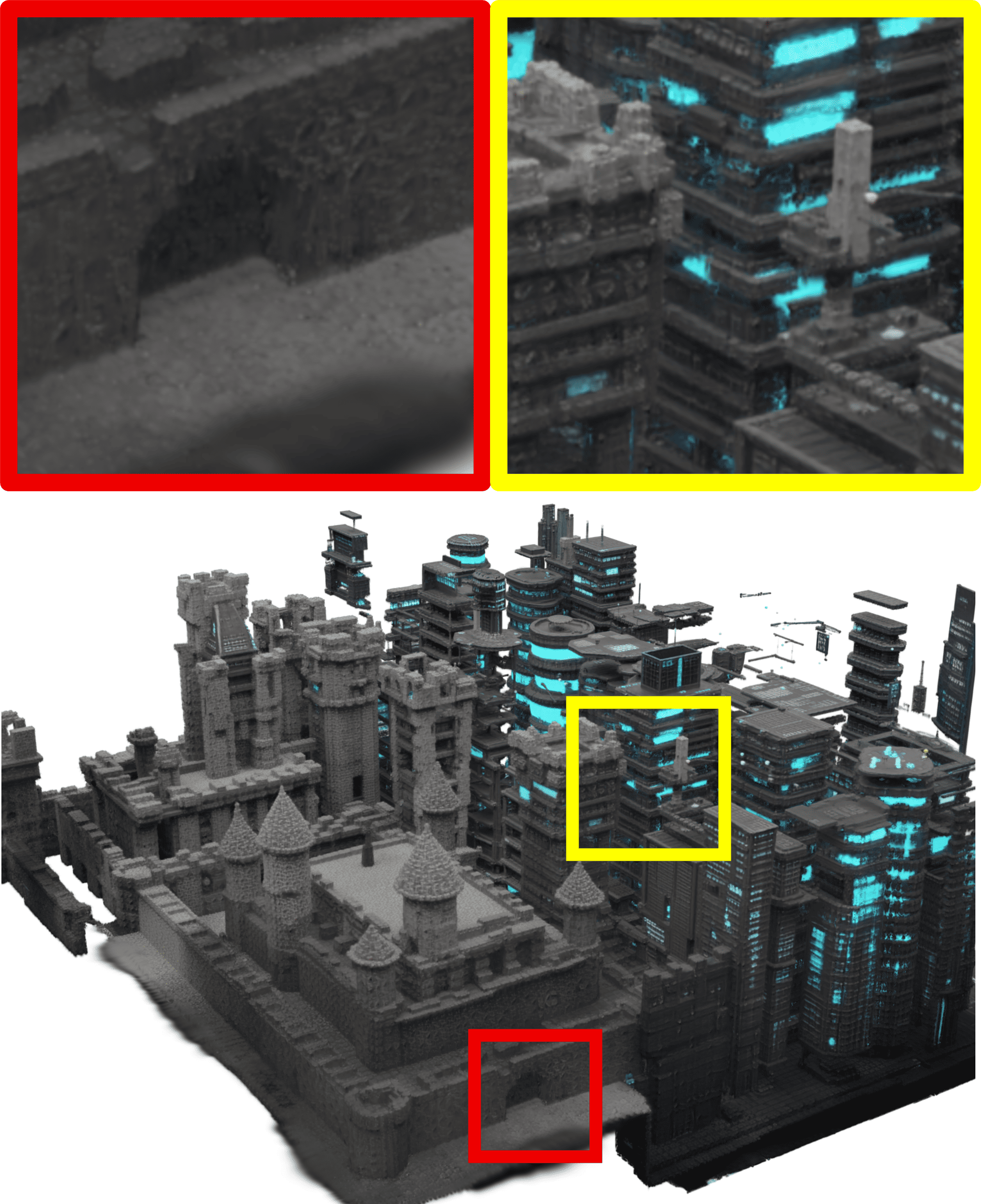}\label{fig:ablation_model_e}}

    \caption{Qualitative comparison of rendered scenes on design choices for the detail enhancer. \textit{Best viewed when zoomed in.}}
    \label{fig:ablation_model}
\end{figure*}

%% file: table/recon_expand.tex
\begin{table}[t]
    \centering
    \caption{
        Quantitative comparison for the detail enhancer design.
        The configuration of our choice and the best metric are written in \textbf{bold}.
    }
    \resizebox{0.8\linewidth}{!}{
    \begin{tabular}{c|ccc|cc|ccc}
        \toprule
        & \multicolumn{2}{c}{$\bm{\mathcal{G}}_{S/L}$} & $\bm{\mathcal{D}}_L$ & \multirow{2}{*}{PSNR$_\uparrow$} & \multirow{2}{*}{LPIPS$_\downarrow$}& \multicolumn{3}{c}{FID$_\downarrow$} \\
        & Architecture & CFG & F.T. & & & Incep.v3 & DINOv2 & CLIP \\
        \midrule
        (a) & \textbf{Concatenation} & \textbf{No} & \textbf{Yes} & \textbf{22.53} & \textbf{0.2137} & \textbf{16.98} & \textbf{32.67} & \textbf{11.79} \\
        (b) & IP-Adapter & No & Yes & 20.28 & 0.2499 & 29.62 & 80.81 & 19.85 \\
        (c) & Concatenation & Yes & Yes & 21.95 & 0.2174 & 19.06 & 38.15 & 21.32 \\
        (d) & Concatenation & No & No & 22.08 & 0.2165 & 17.89 & 32.94 & 13.17 \\
        \bottomrule
    \end{tabular}
    }
    \label{tab:recon}
\end{table}

%% file: sec/5_conclusion.tex
\section{Conclusion} \label{sec:conclusion}
Despite the growing demand for creating worlds, existing methods fail to achieve both flexible and scalable 3D world generation due to the shortage of appropriate data.
In this work, we present Map2World, a novel text-guided 3D world generation pipeline by exploiting the prior knowledge of TRELLIS~\cite{xiang2025structured}, a popular 3D object generation model.
We divide the generation process into two stages.
First, we generate a structured latent that encodes the entire world.
We impose constraints on noise initialization and employ a latent fusion strategy~\cite{bar2023multidiffusion} during sampling to share features across the world to ensure global consistency.
Also, our model supports segment-map-guided sampling and allows users to generate the world from a user-defined map composed of arbitrary-shaped regions, which is not available in any existing works.
Next, we propose a detail enhancing module to further improve the quality of the generated world.
The detail enhancer incorporates the global structure latent as a condition to preserve the consistency of generated details on a global scale.
The detail enhancer is trained by fine-tuning an MLP layer with a small number of parameters that is added prior to the frozen TRELLIS generator, which enables maintaining the generalizable capacity of TRELLIS.
Our model successfully generates worlds that are well aligned with the segment maps and exhibit smooth boundary transitions, even for segment maps of diverse sizes and shapes.


%% file: sec/X_supple.tex
\appendix
\clearpage
\setcounter{page}{1}

\setcounter{section}{0}
\setcounter{figure}{0}
\setcounter{table}{0}

\renewcommand{\thesection}{S\arabic{section}}
\renewcommand{\thetable}{S\arabic{table}}
\renewcommand{\thefigure}{S\arabic{figure}}

\title{\textit{Supplementary Materials for} \\ Map2World: Segment Map Conditioned \\
Text to 3D World Generation
} 
\author{}
\institute{}
\maketitle

\section{Implementation Details}

\subsection{Details on Network Architectures}

\input{supp/fig/slat_interp}
\subsubsection{Feature interpolation for structured latent flow Transformer ($\mathcal{G}_L$).}
The latent tensor used to predict the positions of active voxels($\bm{p}_i$) is a typical 5D tensor with a spatial dimension in the form of a standard 3D cube, and it is straightforward to concatenate with the conditional latent tensors along the channel dimension.
However, when predicting the latent feature ($\bm{z}_i$) for each active voxel, the latent feature is not defined at every position but only at locations where active voxels exist.
Consequently, the conditional latent feature corresponding to a target cube’s active voxel position may be absent.
To address the issue, we estimate the conditional latent feature value at the target position using trilinear interpolation from neighboring positions.
This trilinear interpolation is fully parallelized, enabling fast execution.
To incorporate the latent information from adjacent cubes($\bm{s}^{Adj(j)}$), we expand the spatial size of the target cube.
In this process, the structured latent retains the original positions of the adjacent cube latents. As a result, the latent features of adjacent cubes are concatenated directly without interpolation, and self-attention is then performed on this expanded target cube latent.

The process for concatenating features is illustrated in \cref{fig:supp_slat}.
In the figure, the green and purple quadrangles indicate the spatial axes, with the vertical direction corresponding to the channel axis.
To estimate the features at the positions of the target cube (blue bars), we apply trilinear interpolation from the original positions of the condition latent (orange bars).
For the expanded part of adjacent cubes, we directly bring the features and the positions of adjacent condition latents (pink bars) and concatenate with noises that are expanded at the same positions of condition latents.


\input{supp/table/params}

\subsubsection{Proportion of number of fine-tuned parameters.}
We write the number of fine-tuned parameters and the total parameters in \cref{tab:supp_params}.
The number of fine-tuned parameters is approximately 4\% of the total, similar to the ratio observed in typical LoRA fine-tuning.

\subsection{Training Details}

\subsubsection{Hyperparameter settings.}
We follow the configurations of the original TRELLIS when fine-tuning the flow Transformers and the 3DGS decoder, with 100k iterations for all networks.
The batch size is set to 4 for the geometry flow Transformer($\mathcal{G}_S$), and 8 for the texture flow Transformer($\mathcal{G}_L$) and SLAT decoder($\mathcal{D}_L$).
Each fine-tuning is performed on a single NVIDIA A100 80GB GPU and takes approximately 60 hours per network.

\subsubsection{Fine-tuning detail enhancers.}
As mentioned in the last paragraph of \cref{sec:init_train_samp}, the cubes are sampled in an auto-regressive way during inference.
Thus, the number of adjacent cubes used for the detail enhancer varies from 0 to 3, depending on the index of the cube being rendered.
We design the model to handle a wide range of adjacent cube conditions within a single architecture.
To ensure robust sampling results regardless of the number of adjacent cubes, we vary the number of adjacent cubes used for input conditions when training the detail enhancer.
Specifically, during training, we select an adjacent cube along each of the x(left and right), y(back and forth), and z(up and down) axes relative to the target cube, totally extracting three adjacent cube candidates.
Then, at each iteration, we randomly choose the number of adjacent cubes to use, ranging from 0 to 3, and select the corresponding cubes accordingly for that iteration.
Our model is capable of adding sufficiently fine details using only the latent of the large cube, regardless of the existence of adjacent cube information.
It is noteworthy that when computing the flow and subsequently calculating the loss(\cref{eq:calc_flow,eq:calc_loss}), we exclude the regions containing noise introduced for the adjacent cube condition and perform the computation only for the positions corresponding to the target cube.

\input{supp/fig/flow_loss}

\section{Additional Experiment Results}

\input{supp/fig/supp_clip_compare}
\subsection{Measuring Consistency with Segment Map}
To quantitatively assess the quality of region-specific generation under segment-based conditioning, we compute a CLIP-Score–based alignment metric that measures how well each segmented region corresponds to its associated textual prompt. For a controlled comparison with SynCity, we adopt a grid-aligned generation protocol and report both qualitative and quantitative results in \cref{fig:supp_qual_clip_compare}. As illustrated in \cref{fig:supp_qual_clip_compare}a, our method successfully produces significantly larger and more structurally coherent worlds. We further compute region-wise CLIP-Scores based on top-view renderings, and the aggregated results are shown in \cref{fig:supp_qual_clip_compare}b. The scores are averaged over 50 random seeds, capturing the relative alignment between the four segmented regions and their four textual descriptions. Using the ViT-H/14 CLIP model, our approach exhibits substantially clearer region–text separability compared to SynCity, indicating stronger adherence to the intended semantic layout.

Additional grid-wise samples and extended quantitative results, including CLIP-Scores obtained from alternative CLIP backbones (i.e., ViT-L/14, ViT-bigG/14, SigLIP-So400m/14, PE-Core/14) as well as their softmax-normalized variants, are presented in \cref{fig:supp_qual_clip_grid}. These results demonstrate that our method maintains consistent region-text alignment across different backbone configurations.

Furthermore, \cref{fig:supp_qual_clip_free1,fig:supp_qual_clip_free2,fig:supp_qual_clip_free3} show that our framework generalizes effectively to arbitrary user-defined segmentation maps. 
Given a free-form segmentation mask and textual descriptions, the proposed method reliably generates 3D scenes whose spatial structure and semantic content faithfully reflect the provided user inputs, highlighting the flexibility and robustness of our approach.
For completeness, the quantitative results presented in these figures are aggregated over 50 randomly seeded generations, ensuring a consistent and statistically stable comparison across different CLIP backbones and scene configurations.
Each figure shows representative samples and their averaged CLIP-Scores.

\subsection{Detail Enhancer}

\subsubsection{Evaluation protocol.}
Since the evaluation metrics require ground truth images, we used 1,500 cubes designated as test samples during Objaverse data curation. For each cube, we generated four images by rotating the yaw angle by 90° increments while maintaining a radius of 2 and looking toward the cube’s center, with a field of view (FoV) of 40°. This resulted in a total of 6,000 images for evaluation.



\subsubsection{Recursive detail enhancement.}
Since the detail enhancer is trained by data with various metric scales, we can recursively apply the detail enhancer to further upscale the resolution of the scene.
We visualize the rendered images of recursively enhanced worlds in \cref{fig:supp_recursive}.
To evaluate the performance of the detail enhancer alone, we first generate the initial structured latent without applying any latent fusion strategy, producing a cube of size 64 as in the original TRELLIS output, which is displayed in \cref{fig:supp_recursive_x1}.
This cube is then progressively upscaled by factors of 2 (\cref{fig:supp_recursive_x2}) and 4 (\cref{fig:supp_recursive_x4}) through the detail enhancer.

The top row shows the example from an urban scene from the Objaverse dataset.
Here, the image in the left column is rendered by sequentially encoding and decoding the input 3D scene.
The windows, before passing through the detail enhancer, are blurred, but their boundaries become sharper after passing through the enhancer.
The second and third rows are rendered from TRELLIS-generated samples.
Due to the output size limit of the original model, the images in the left column appear blurred and cannot represent fine-grained textures.
Although the detail enhancer slightly transforms the contents of the original 3D world ($\times1$), it generates sharper details while maintaining harmony with surrounding elements.
Since our goal is not to perfectly reconstruct the original 3D world but to produce a high-quality world aligned with the input text prompt, we argue that minor content modifications are acceptable if they enhance overall quality.

\subsection{Additional Qualitative Results}
In \cref{fig:supp_qual_freeform}, we presented results obtained through latent fusion and the detail enhancer for a wider variety of segment maps.
Our model demonstrates strong adherence to the given conditions and produces high-quality outputs even for irregularly shaped segment maps that SynCity cannot handle.

We also attached a video demo that displays the rendered views of the generated world from the given image segment maps.
The video demonstrates that our model successfully generates a high-quality 3D world that faithfully satisfies the segment map condition.

\input{supp/fig/supp_clip_gridseg}
\input{supp/fig/supp_clip_freeseg1}
\input{supp/fig/supp_clip_freeseg2}
\input{supp/fig/supp_clip_freeseg3}

\section{Discussion}
\subsubsection{Limitations.}
Since our model is built on TRELLIS, Map2World share the same limitations of TRELLIS.
First, TRELLIS uses absolute position encoding to give positional information to the model.
However, when applying TRELLIS in our pipeline, merging small cubes can alter positional information before and after the merge, which may lead to changes in the decoded 3D structure.
This issue can be mitigated by using a baseline model that uses relative positions instead of absolute positions, or by applying training strategies that enable the model to adapt effectively to changing positional encodings, thereby improving the overall quality of the results.

\subsubsection{Future works.}
While the current detail enhancer is fine-tuned only with scene-level cropped data, the generalizability of the enhancer can be improved by using more data.
Specifically, quality can be improved by training on both object-level and world-level data.
Additionally, since Objaverse primarily consists of simple meshes, incorporating datasets with more complex and realistic textures would encourage the detail enhancer to move toward a more photo-realistic direction.

\subsubsection{Social impact.}
Our work focuses on controllable 3D scene generation and does not directly address personal data, identity modeling, or downstream decision-making tasks. Accordingly, we do not anticipate significant adverse social impacts. Potential applications—such as simulation, virtual environment creation, and content generation—are primarily creative or industrial, and the method does not inherently facilitate misuse beyond the general considerations associated with generative models. We encourage responsible use within appropriate ethical and safety guidelines.

\input{supp/fig/X4SR}
\input{supp/fig/supp_qual}

%% file: supp/fig/slat_interp.tex
\begin{figure}[t]
    \centering
    \includegraphics[width=0.6\linewidth]{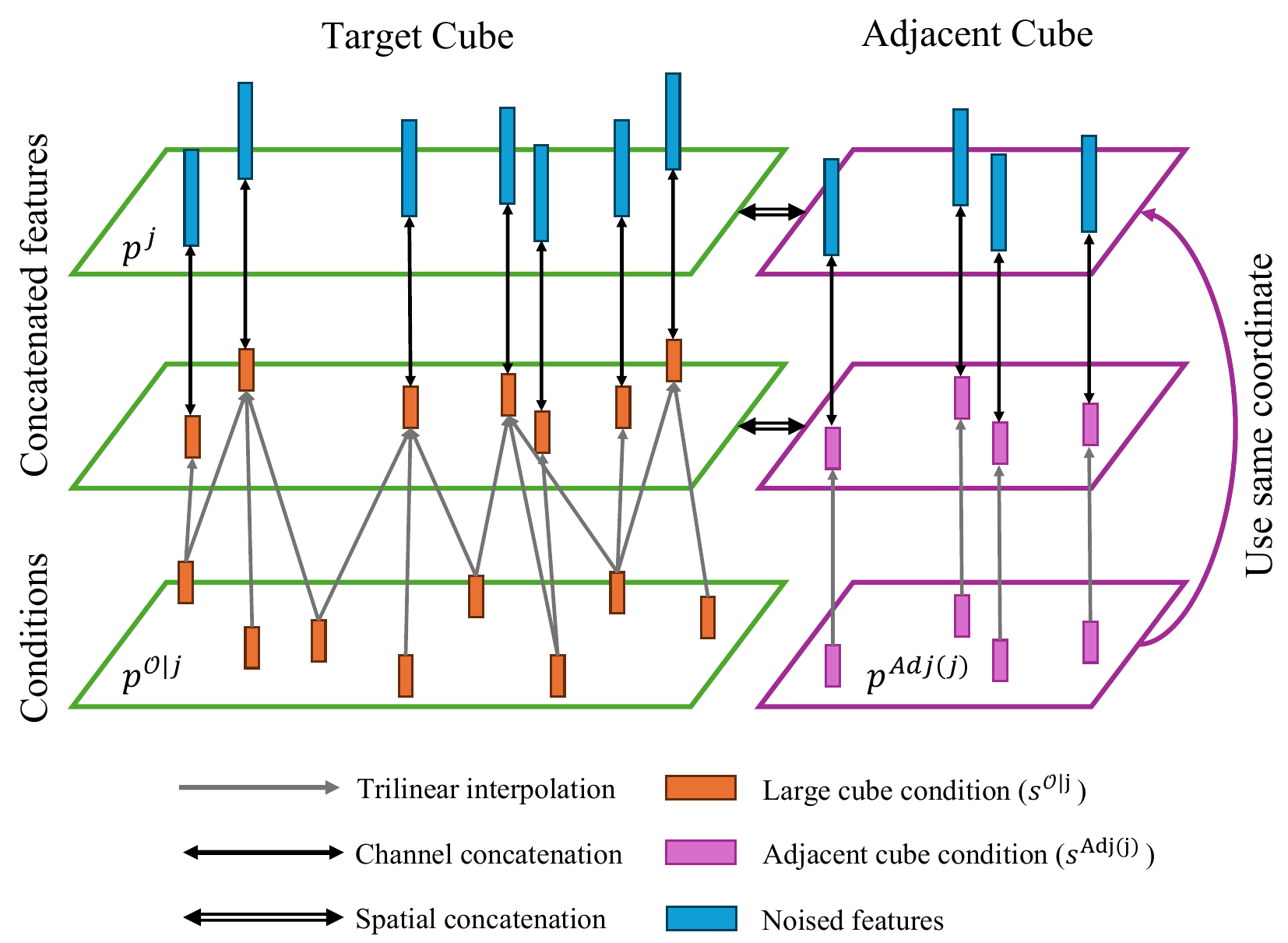}
    \captionof{figure}{
        Visualization of concatenating feature process in structured latent flow Transformer ($\mathcal{G}_L$).
    }
    \label{fig:supp_slat}
\end{figure}

%% file: supp/table/params.tex
\begin{table}[t]
    \centering
    \caption{Number of trained parameters and total parameters for each flow Transformer.}
    \resizebox{0.6\linewidth}{!}{
    \begin{tabular}{c|cc|c}
        \toprule
        \multirow{2}{*}{Base network} & \multicolumn{2}{c|}{\# of parameters} & Fine-tuned \\
         & Fine-tuned & Total & Percent(\%) \\
        \midrule
        $\mathcal{G}_S$ & 25.39M & 580.93M & 4.37\% \\
        $\mathcal{G}_L$ & 25.39M & 625.82M & 4.06\% \\
        \bottomrule
    \end{tabular}
    }
    \label{tab:supp_params}
    \vspace{-2mm}
\end{table}

%% file: supp/fig/flow_loss.tex
\begin{figure}[t]
    \centering
    \includegraphics[width=0.4\linewidth]{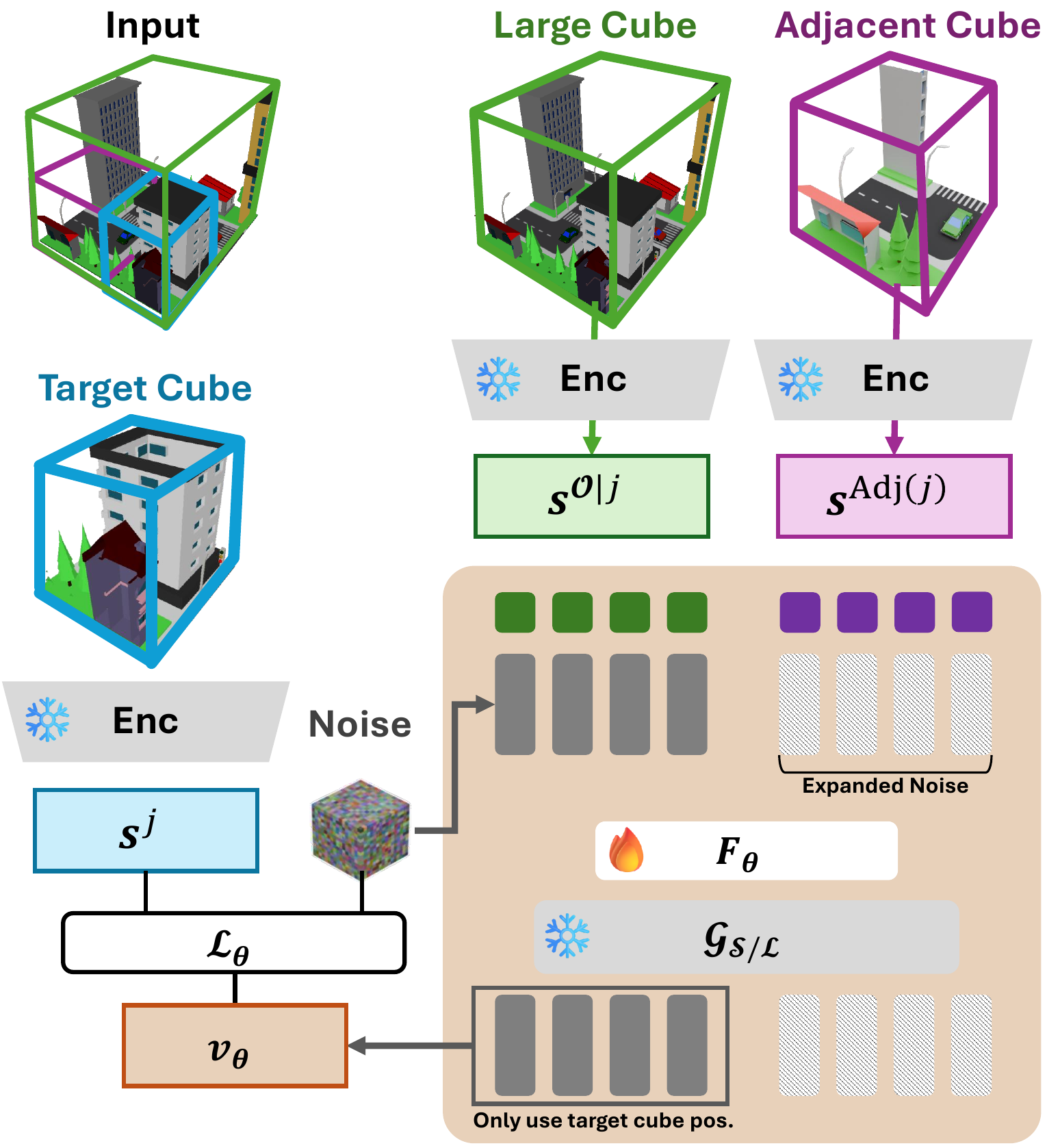}
    \captionof{figure}{
        Visualization of the estimation of the flow tensor and the calculation of the loss.
    }
    \label{fig:supp_loss}
\end{figure}

%% file: supp/fig/supp_clip_compare.tex
\begin{figure}[t]
    \centering
    \subfloat{\includegraphics[width=0.5\linewidth]{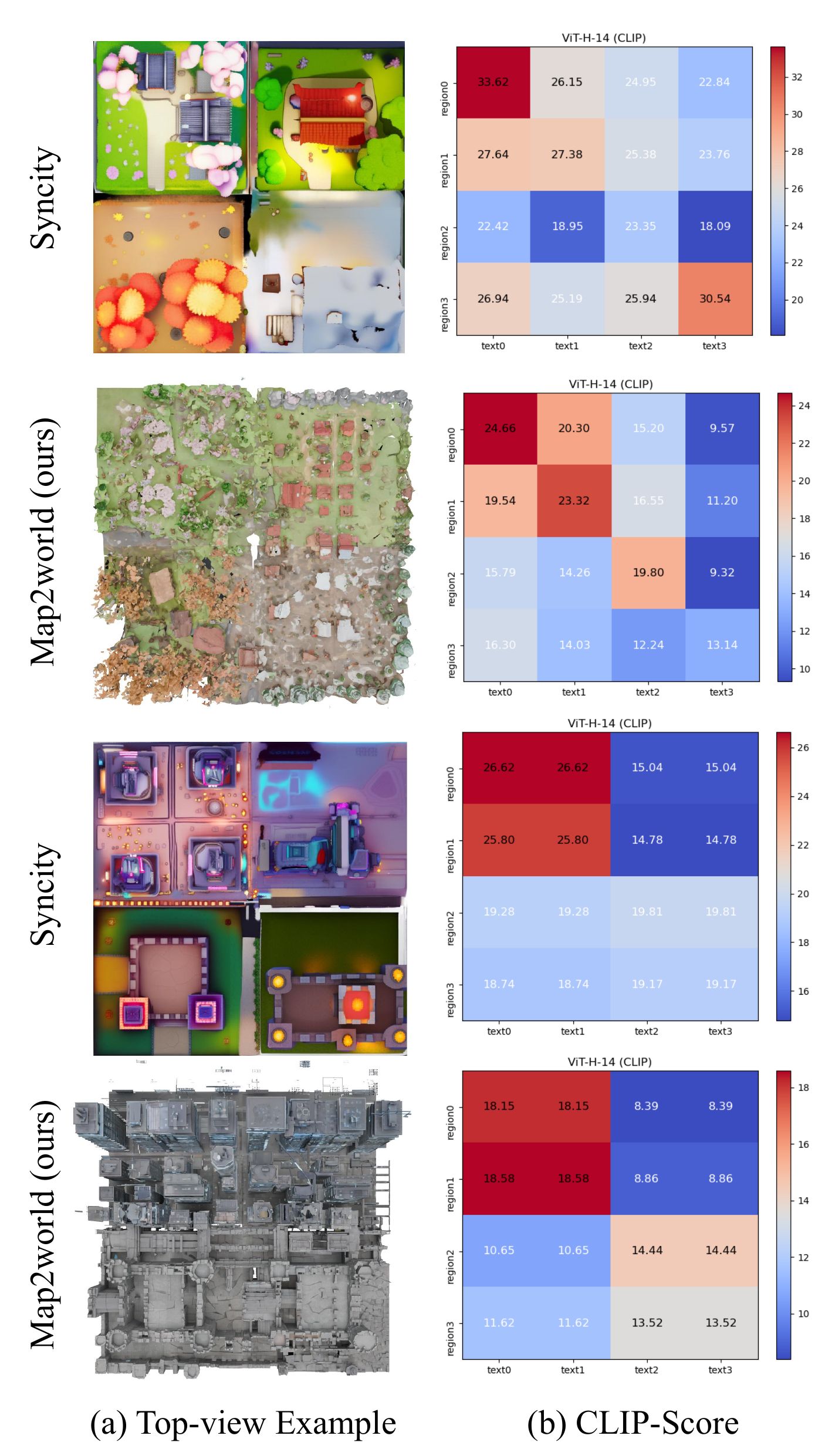}}
    \caption{
        Comparisons in CLIP-Score heatmap evaluated with ViT-H-14 baseline.
    }
    \label{fig:supp_qual_clip_compare}
\end{figure}

%% file: supp/fig/supp_clip_gridseg.tex
\begin{figure*}[t]
    \newcommand{\ww}{\linewidth}
    \centering

    \subfloat{\includegraphics[width=0.8\ww]{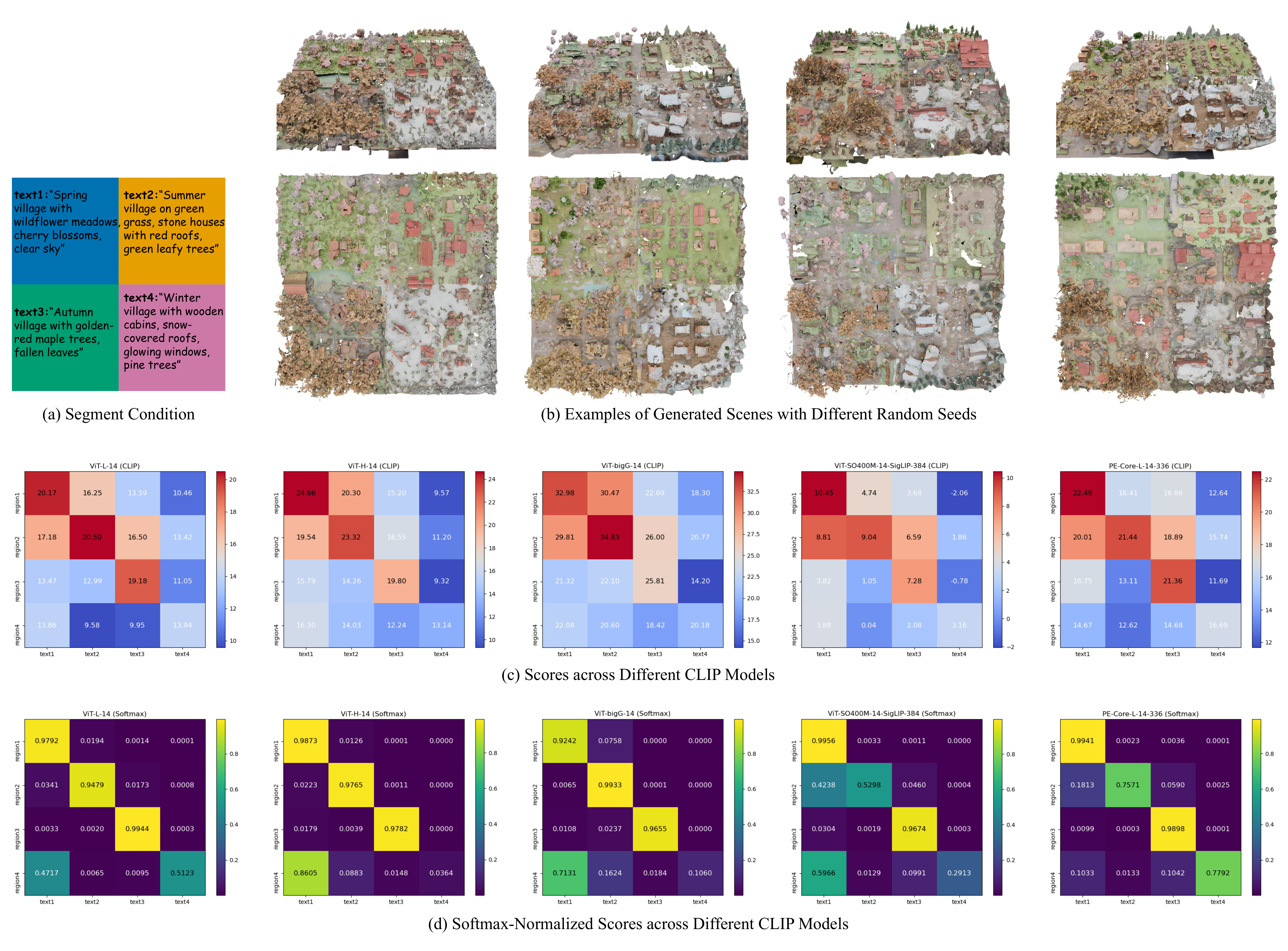}}

    \vspace{3mm}
    
    \subfloat{\includegraphics[width=0.8\ww]{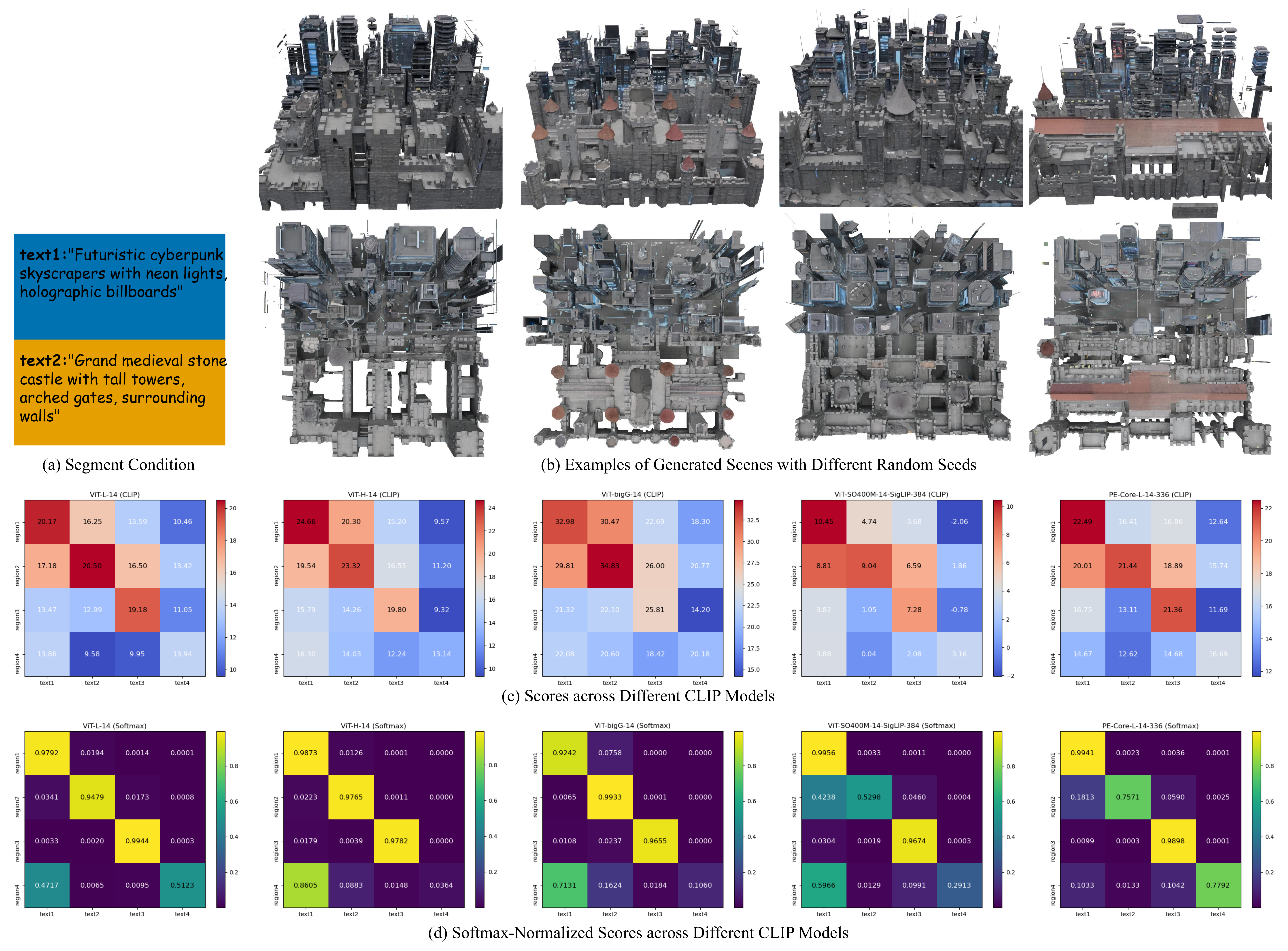}}
    
    \caption{
        CLIP-Score heatmap visualization with grid-type segmentation map conditions.
        \textit{Best viewed when zoomed in.}
    }
    \label{fig:supp_qual_clip_grid}
\end{figure*}

%% file: supp/fig/supp_clip_freeseg1.tex
\begin{figure*}[t]
    \newcommand{\ww}{\linewidth}
    \centering

    \subfloat{\includegraphics[width=\ww]{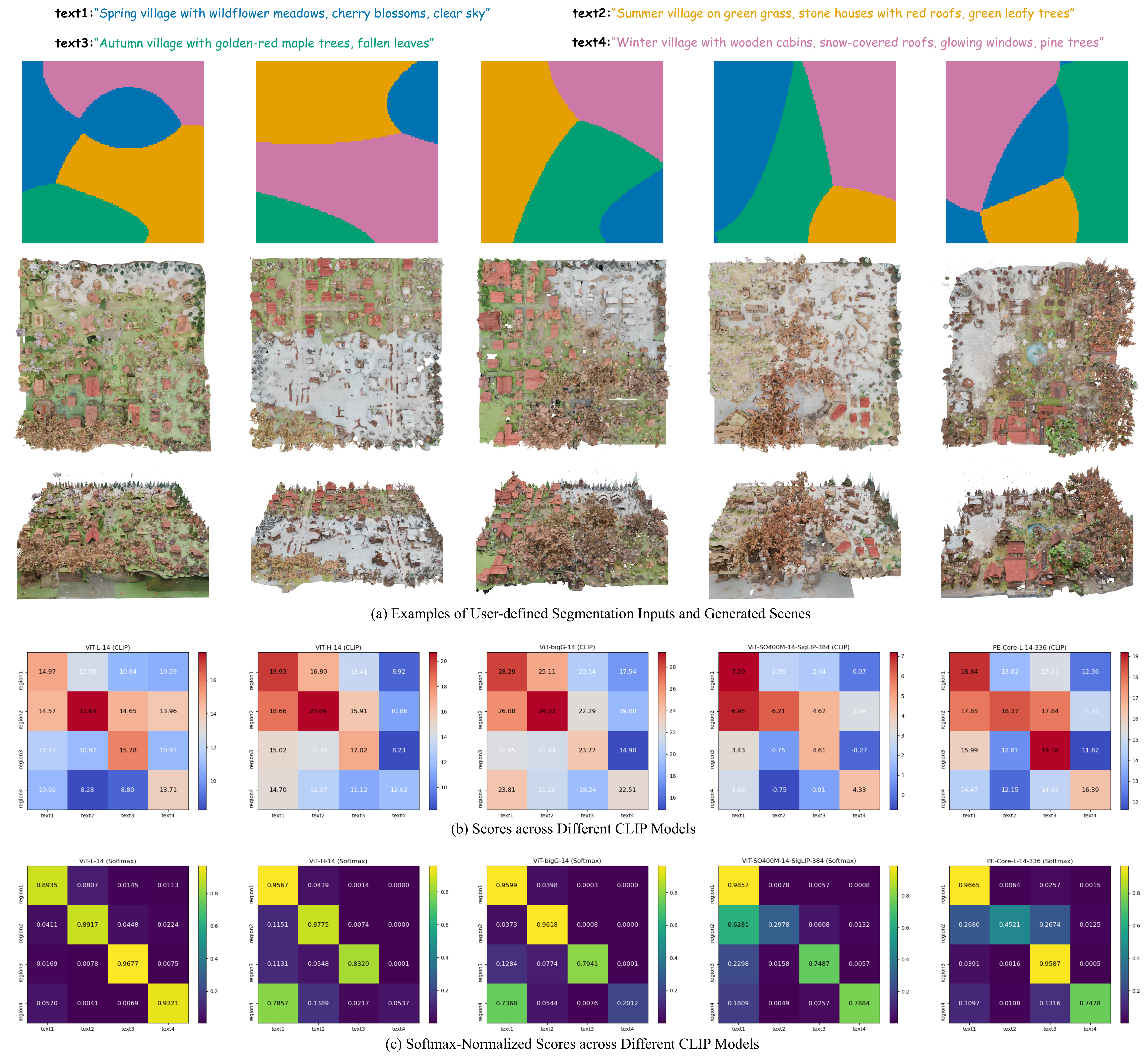}}
    
    \caption{
        CLIP-Score heatmap visualization with arbitrary-shape segmentation map conditions.
        \textit{Best viewed when zoomed in.}
    }
    \label{fig:supp_qual_clip_free1}
\end{figure*}

%% file: supp/fig/supp_clip_freeseg2.tex
\begin{figure*}[t]
    \newcommand{\ww}{\linewidth}
    \centering

    \subfloat{\includegraphics[width=\ww]{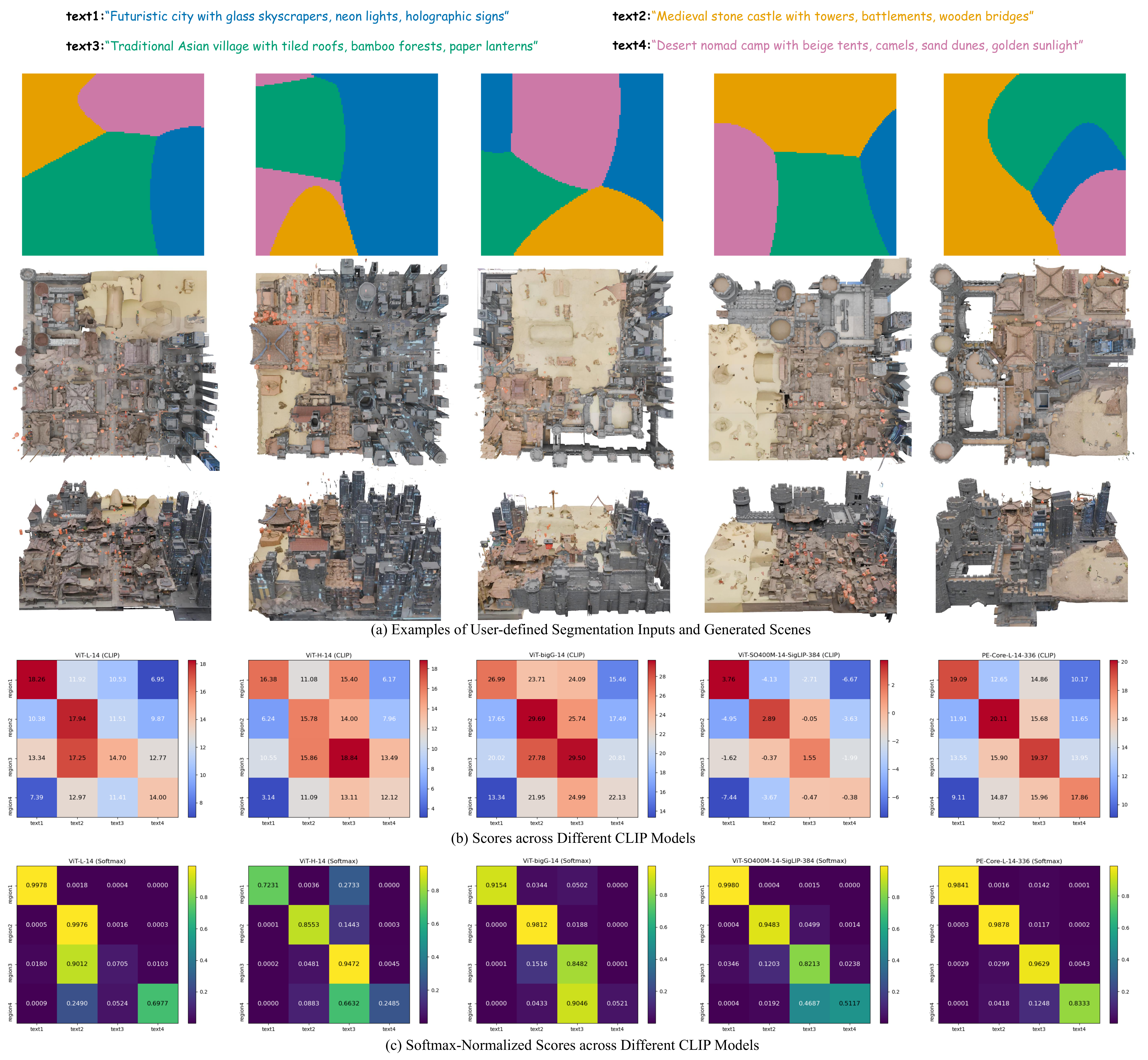}}
    
    \caption{
        CLIP-Score heatmap visualization with arbitrary-shape segmentation map conditions.
        \textit{Best viewed when zoomed in.}
    }
    \label{fig:supp_qual_clip_free2}
\end{figure*}

%% file: supp/fig/supp_clip_freeseg3.tex
\begin{figure*}[t]
    \newcommand{\ww}{\linewidth}
    \centering

    \subfloat{\includegraphics[width=\ww]{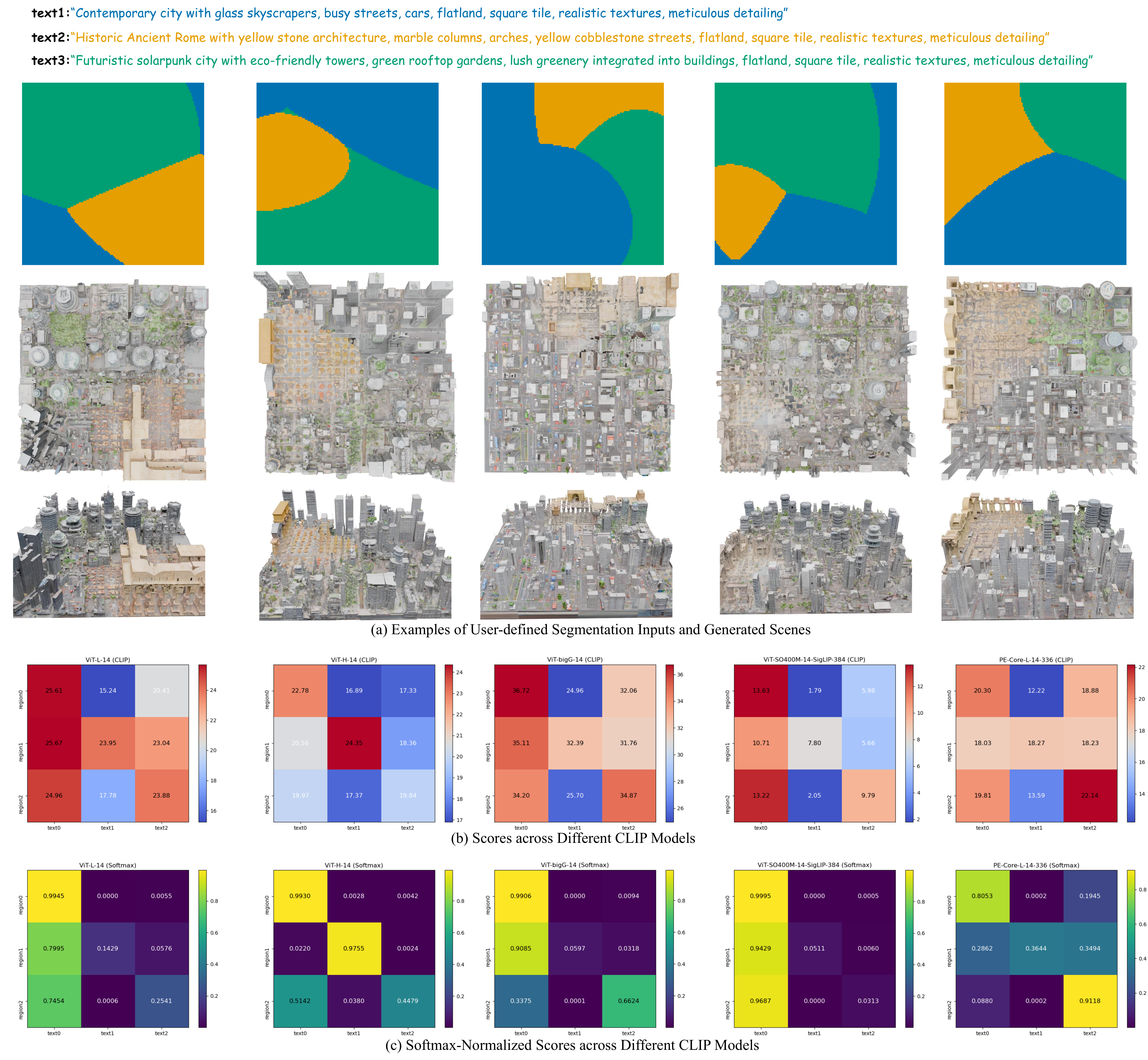}}
    
    \caption{
        CLIP-Score heatmap visualization with arbitrary-shape segmentation map conditions.
        \textit{Best viewed when zoomed in.}
    }
    \label{fig:supp_qual_clip_free3}
\end{figure*}

%% file: supp/fig/X4SR.tex
\begin{figure*}[t]
    \newcommand{\ww}{0.32\linewidth}
    \centering

        

    \subfloat[$\times1$\label{fig:supp_recursive_x1}]{\includegraphics[width=\ww]{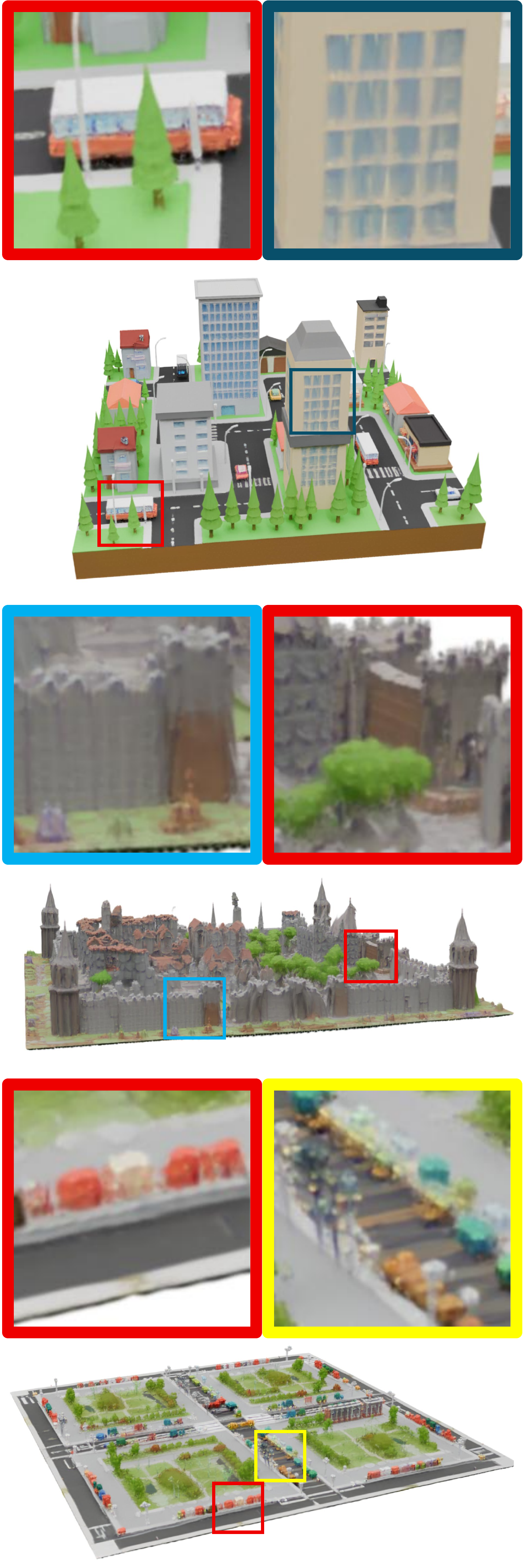}}
    \hfill
    \subfloat[$\times2$\label{fig:supp_recursive_x2}]{\includegraphics[width=\ww]{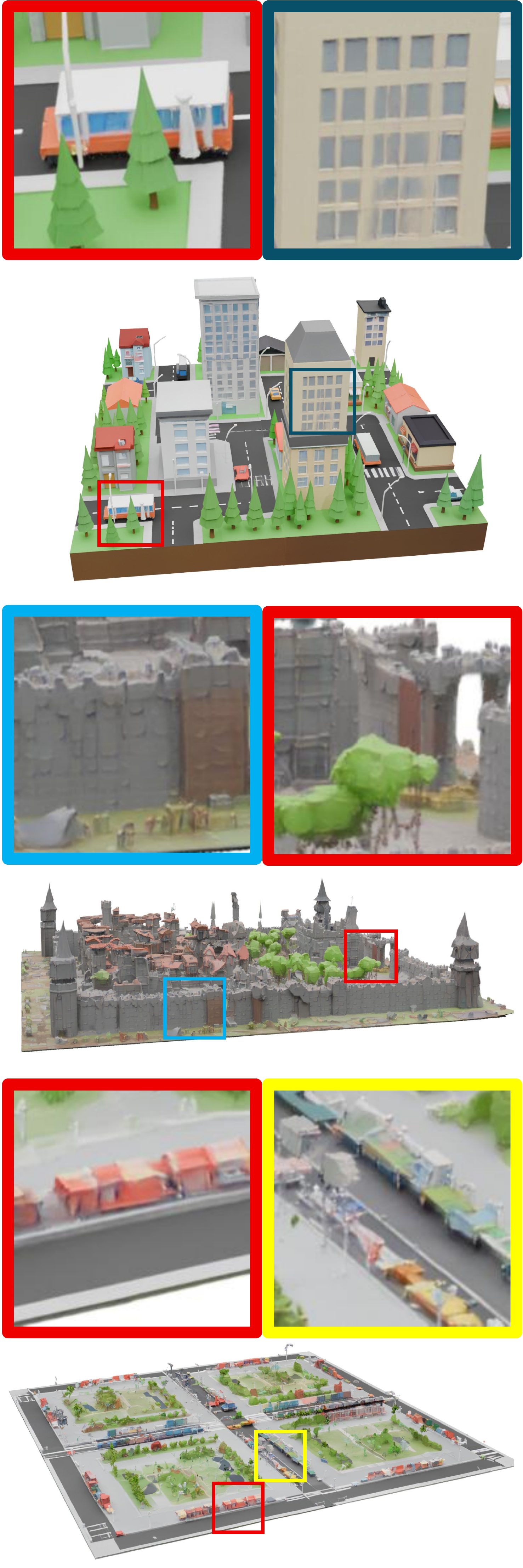}}
    \hfill
    \subfloat[$\times4$\label{fig:supp_recursive_x4}]{\includegraphics[width=\ww]{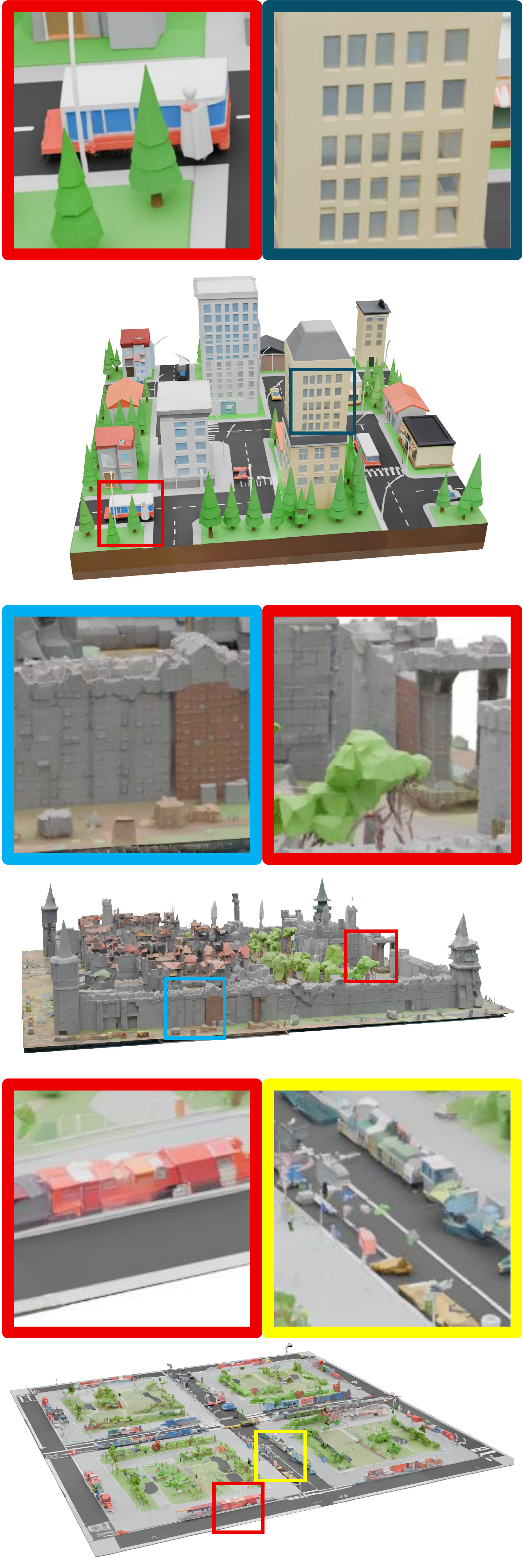}}
    
    \caption{
        Qualitative comparison of recursive detail enhancement.
        We do not use latent fusion tricks when generating the initial world ($\times1$) in this experiment.
        $\times4$ indicates the latent tensors are passed through the detail enhancer twice to further upscale the resolution of the world.
    }
    \label{fig:supp_recursive}
\end{figure*}

%% file: supp/fig/supp_qual.tex
\begin{figure*}[t]
    \newcommand{\w}{0.175\linewidth}
    \newcommand{\ww}{0.35\linewidth}
    \newcommand{\www}{0.28\linewidth}
    \newcommand{\hh}{0.175\linewidth}
    \centering

    \subfloat{\includegraphics[width=\ww, height=\hh]{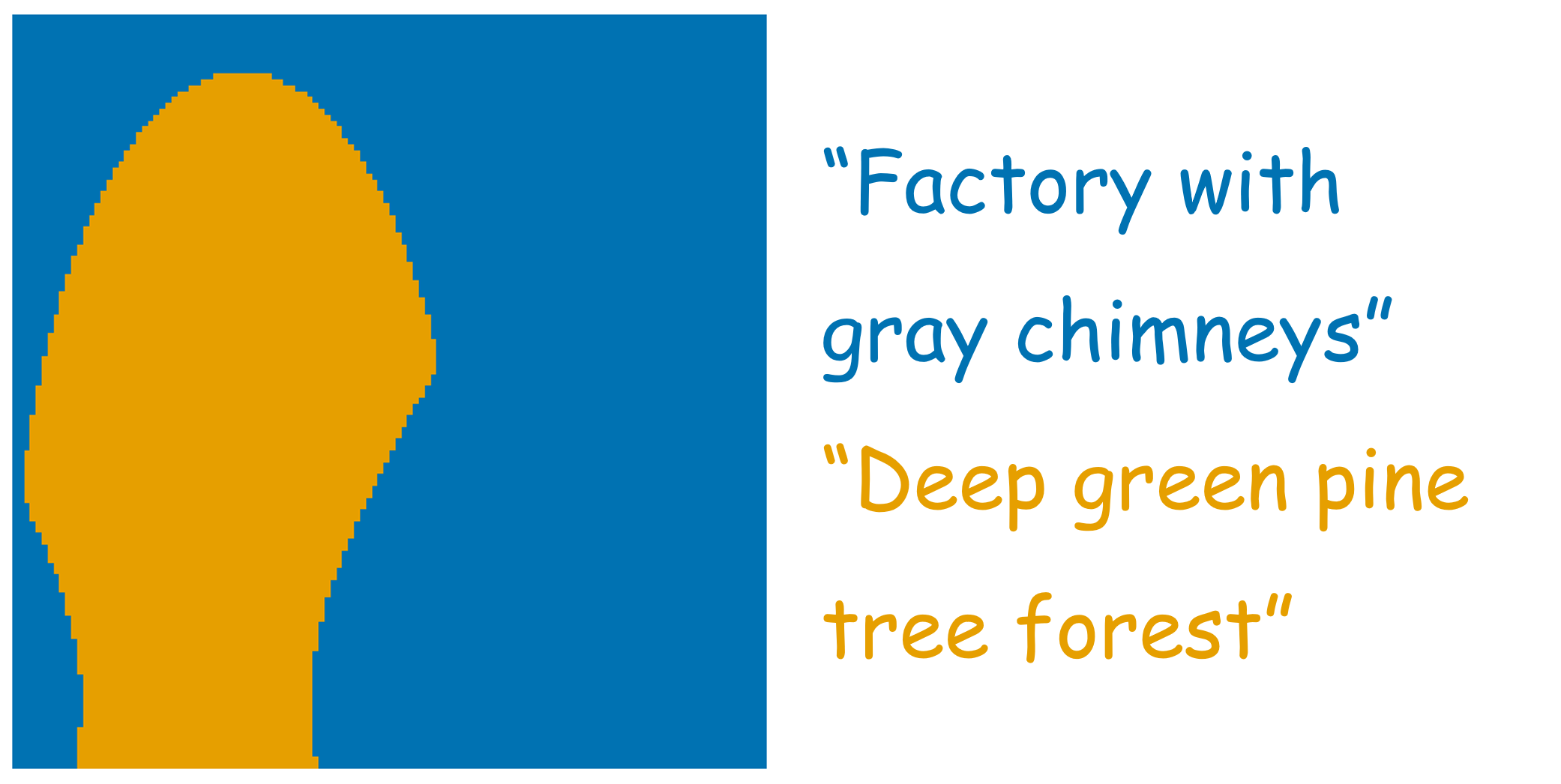}}
    \hfill
    \subfloat{\includegraphics[width=\w, height=\hh]{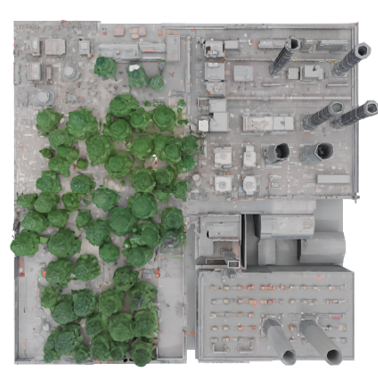}}
    \hfill
    \subfloat{\includegraphics[width=\www, height=\hh]{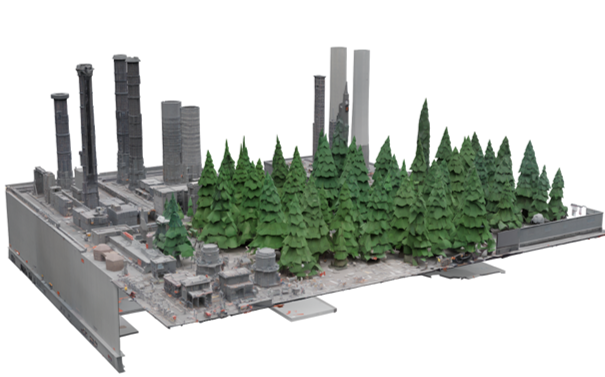}}
    \hfill
    \subfloat{\includegraphics[width=\w, height=\hh]{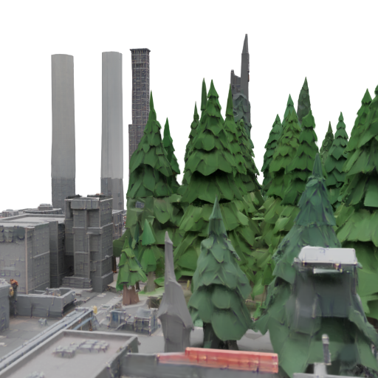}}

    \vspace{3mm}
    
    \addtocounter{subfigure}{-4}
    \subfloat[Segment map]{\includegraphics[width=\ww]{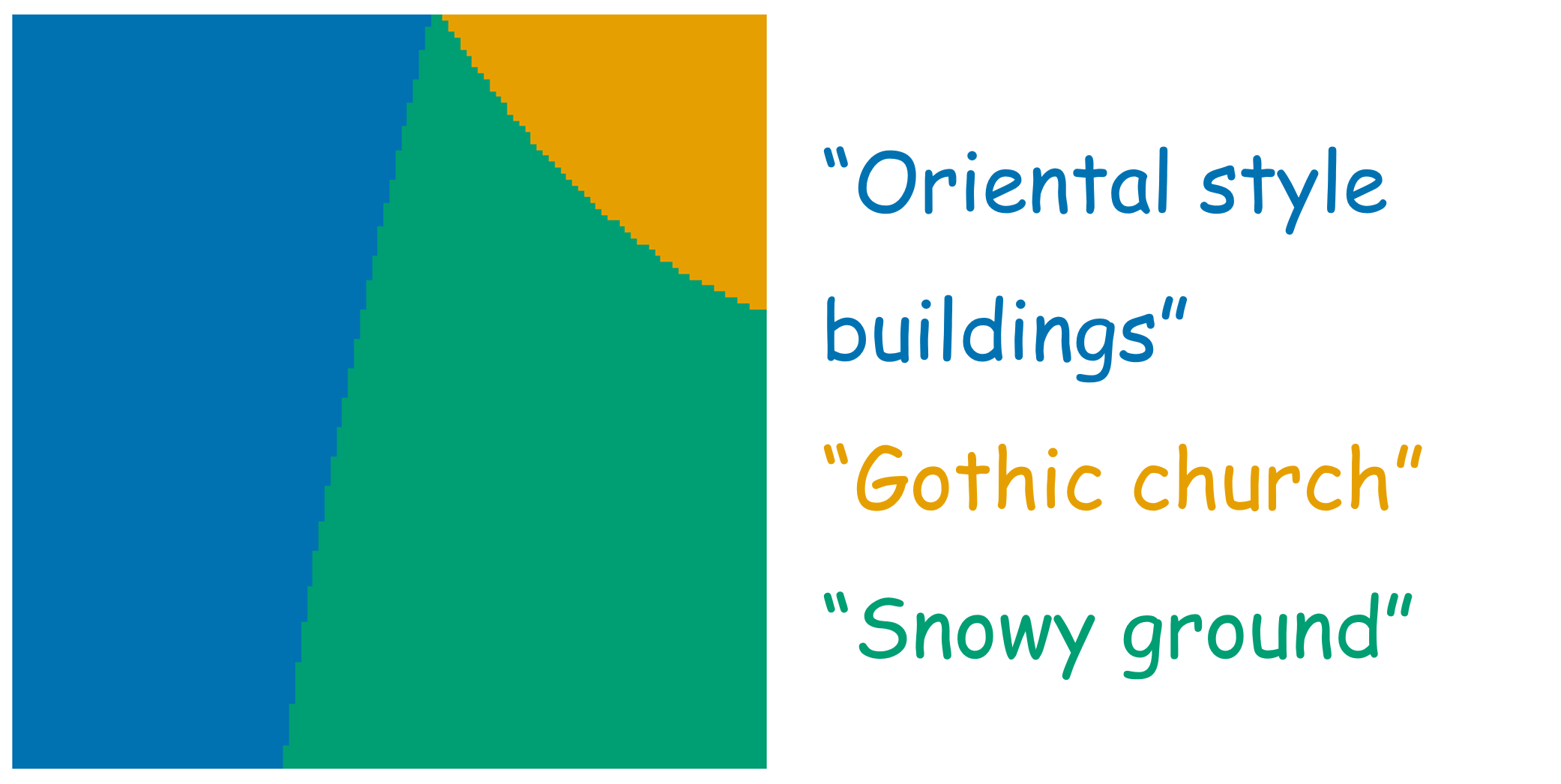}} 
    \hfill
    \subfloat[Top-view map]{\includegraphics[width=\w]{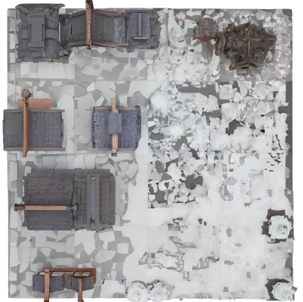}} 
    \hfill
    \subfloat[Generated world]{\includegraphics[width=\www]{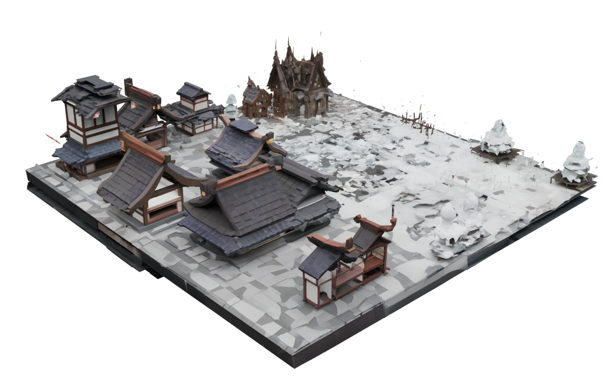}}
    \hfill
    \subfloat[Roaming view]{\includegraphics[width=\w]{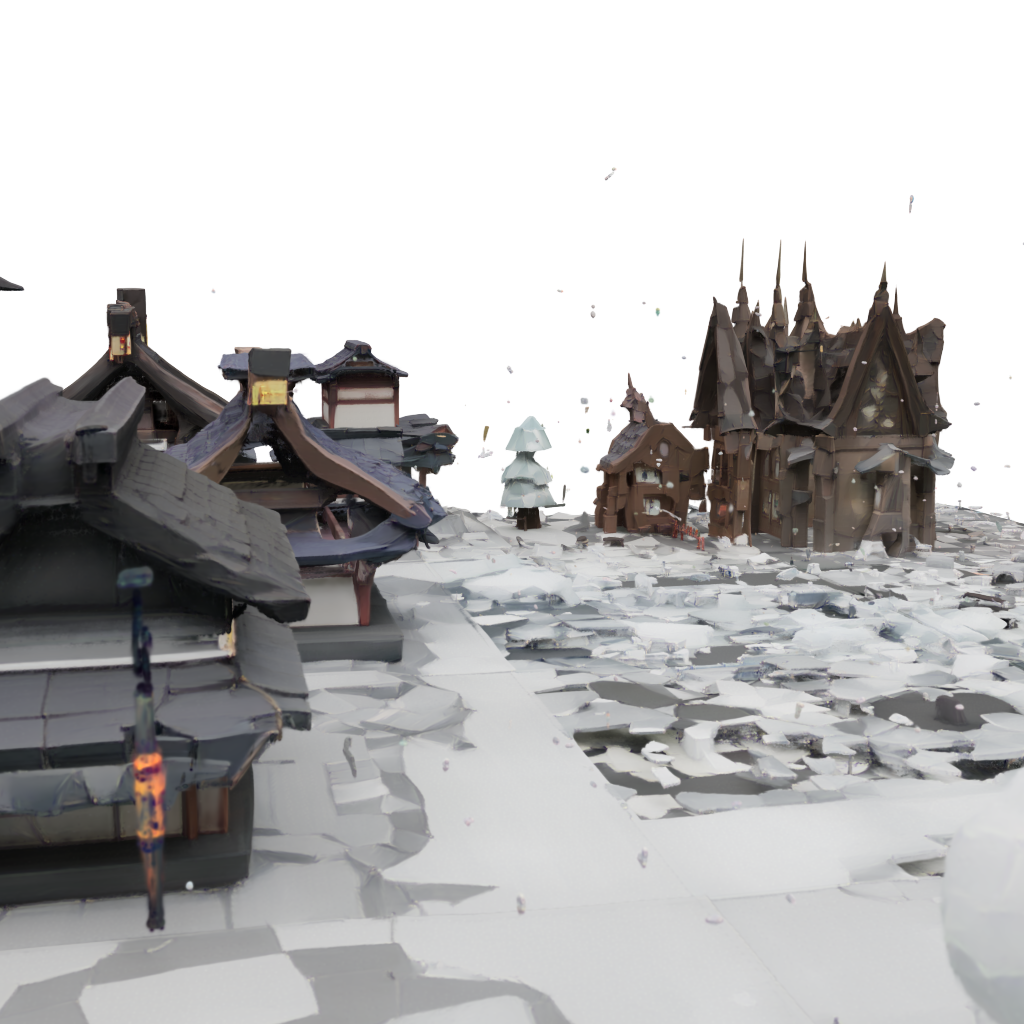}}
    
    \caption{
        Additional qualitative results conditioned by arbitrarily shaped segment maps with user-defined text prompts.
        Our model creates a 3D world that corresponds to the input segment map regardless of the shape of each region.
        We stress that \textbf{SynCity cannot generate the scene from such a complicated segment map}.
    }
    \label{fig:supp_qual_freeform}
\end{figure*}

%% file: main.bib
@String(CVPR  = {IEEE Conf. Comput. Vis. Pattern Recog.})

@String(ICCV  = {Int. Conf. Comput. Vis.})

@String(NeurIPS = {Adv. Neural Inform. Process. Syst.})

@String(TOG   = {ACM Trans. Graph.})

@String(CVPR  = {CVPR})

@String(ICCV  = {ICCV})

@String(NeurIPS = {NeurIPS})

@String(TOG   = {ACM TOG})

@inproceedings{xiang2025structured,
    title={Structured 3d latents for scalable and versatile 3d generation},
    author={Xiang, Jianfeng and Lv, Zelong and Xu, Sicheng and Deng, Yu and Wang, Ruicheng and Zhang, Bowen and Chen, Dong and Tong, Xin and Yang, Jiaolong},
    booktitle={CVPR},
    pages={21469--21480},
    year={2025}
}

@article{kerbl20233d,
    title={3D Gaussian splatting for real-time radiance field rendering.},
    author={Kerbl, Bernhard and Kopanas, Georgios and Leimk{\"u}hler, Thomas and Drettakis, George},
    journal={ACM Trans. Graph.},
    volume={42},
    number={4},
    pages={139--1},
    year={2023}
}

@inproceedings{gao2023strivec,
    title={Strivec: Sparse tri-vector radiance fields},
    author={Gao, Quankai and Xu, Qiangeng and Su, Hao and Neumann, Ulrich and Xu, Zexiang},
    booktitle={Proceedings of the IEEE/CVF International Conference on Computer Vision},
    pages={17569--17579},
    year={2023}
}

@article{shen2023flexible,
    title={Flexible isosurface extraction for gradient-based mesh optimization},
    author={Shen, Tianchang and Munkberg, Jacob and Hasselgren, Jon and Yin, Kangxue and Wang, Zian and Chen, Wenzheng and Gojcic, Zan and Fidler, Sanja and Sharp, Nicholas and Gao, Jun},
    journal={ACM Transactions on Graphics (TOG)},
    volume={42},
    number={4},
    pages={1--16},
    year={2023},
    publisher={ACM New York, NY, USA}
}

@article{wu2024blockfusion,
    title={BlockFusion: Expandable 3D Scene Generation using Latent Tri-plane Extrapolation},
    author={Wu, Zhennan and Li, Yang and Yan, Han and Shang, Taizhang and Sun, Weixuan and Wang, Senbo and Cui, Ruikai and Liu, Weizhe and Sato, Hiroyuki and Li, Hongdong and Ji, Pan},
    journal={ACM TOG},
    year={2024},
}

@article{lee2025nuiscene,
    title={NuiScene: Exploring efficient generation of unbounded outdoor scenes},
    author={Lee, Han-Hung and Han, Qinghong and Chang, Angel X},
    journal={arXiv preprint arXiv:2503.16375},
    year={2025}
}

@inproceedings{meng2025lt3sd,
    title={Lt3sd: Latent trees for 3d scene diffusion},
    author={Meng, Quan and Li, Lei and Nie{\ss}ner, Matthias and Dai, Angela},
    booktitle={Proceedings of the Computer Vision and Pattern Recognition Conference},
    pages={650--660},
    year={2025}
}

@inproceedings{ren2024xcube,
    title={XCube: Large-Scale 3D Generative Modeling using Sparse Voxel Hierarchies}, 
    author={Ren, Xuanchi and Huang, Jiahui and Zeng, Xiaohui and Museth, Ken 
      and Fidler, Sanja and Williams, Francis},
    booktitle={Proceedings of the IEEE/CVF Conference on Computer Vision and Pattern Recognition},
    year={2024}
}

@inproceedings{ren2024scube,
    title={SCube: Instant Large-Scale Scene Reconstruction using VoxSplats},
    author={Ren, Xuanchi and Lu, Yifan and Liang, Hanxue and Wu, Jay Zhangjie and 
    Ling, Huan and Chen, Mike and Fidler, Sanja and Williams, Francis and Huang, Jiahui},
    booktitle={The Thirty-eighth Annual Conference on Neural Information Processing Systems},
    year={2024},
}

@article{worldgrow2025,
  title   = {WorldGrow: Generating Infinite 3D World},
  author  = {Li, Sikuang and Yang, Chen and Fang, Jiemin and Yi, Taoran and Lu, Jia and Cen, Jiazhong and Xie, Lingxi and Shen, Wei and Tian, Qi},
  journal = {arXiv preprint arXiv:2510.21682},
  year    = {2025}
}

@misc{lu2024infinicube,
    title={InfiniCube: Unbounded and Controllable Dynamic 3D Driving Scene Generation with World-Guided Video 
      Models}, 
    author={Yifan Lu and Xuanchi Ren and Jiawei Yang and Tianchang Shen and Zhangjie Wu and Jun Gao and 
      Yue Wang and Siheng Chen and Mike Chen and Sanja Fidler and Jiahui Huang},
    year={2024},
    eprint={2412.03934},
    archivePrefix={arXiv},
    primaryClass={cs.CV},
    url={https://arxiv.org/abs/2412.03934}, 
}

@InProceedings{hoellein2023text2room,
    author    = {H\"ollein, Lukas and Cao, Ang and Owens, Andrew and Johnson, Justin and Nie{\ss}ner, Matthias},
    title     = {Text2Room: Extracting Textured 3D Meshes from 2D Text-to-Image Models},
    booktitle = {Proceedings of the IEEE/CVF International Conference on Computer Vision (ICCV)},
    month     = {October},
    year      = {2023},
    pages     = {7909-7920}
}

@article{chung2023luciddreamer,
    title={LucidDreamer: Domain-free Generation of 3D Gaussian Splatting Scenes},
    author={Chung, Jaeyoung and Lee, Suyoung and Nam, Hyeongjin and Lee, Jaerin and Lee, Kyoung Mu},
    journal={arXiv preprint arXiv:2311.13384},
    year={2023}
}

@inproceedings{yu2024wonderjourney,
  title={Wonderjourney: Going from anywhere to everywhere},
  author={Yu, Hong-Xing and Duan, Haoyi and Hur, Junhwa and Sargent, Kyle and Rubinstein, Michael and Freeman, William T and Cole, Forrester and Sun, Deqing and Snavely, Noah and Wu, Jiajun and others},
  booktitle={Proceedings of the IEEE/CVF Conference on Computer Vision and Pattern Recognition},
  pages={6658--6667},
  year={2024}
}

@inproceedings{yu2025wonderworld,
  title={Wonderworld: Interactive 3d scene generation from a single image},
  author={Yu, Hong-Xing and Duan, Haoyi and Herrmann, Charles and Freeman, William T and Wu, Jiajun},
  booktitle={Proceedings of the Computer Vision and Pattern Recognition Conference},
  pages={5916--5926},
  year={2025}
}

@inproceedings{shriram2024realmdreamer,
    title={RealmDreamer: Text-Driven 3D Scene Generation with 
            Inpainting and Depth Diffusion},
    author={Jaidev Shriram and Alex Trevithick and Lingjie Liu and Ravi Ramamoorthi},
    booktitle={International Conference on 3D Vision (3DV)},
    year={2025}
}

@inproceedings{li2024dreamscene,
    title={Dreamscene: 3d gaussian-based text-to-3d scene generation via formation pattern sampling},
    author={Li, Haoran and Shi, Haolin and Zhang, Wenli and Wu, Wenjun and Liao, Yong and Wang, Lin and Lee, Lik-hang and Zhou, Peng Yuan},
    booktitle={European Conference on Computer Vision},
    pages={214--230},
    year={2024},
    organization={Springer}
}

@article{liu20243dgs,
  title={3dgs-enhancer: Enhancing unbounded 3d gaussian splatting with view-consistent 2d diffusion priors},
  author={Liu, Xi and Zhou, Chaoyi and Huang, Siyu},
  journal={Advances in Neural Information Processing Systems},
  volume={37},
  pages={133305--133327},
  year={2024}
}

@inproceedings{wang2025videoscene,
    title={VideoScene: Distilling video diffusion model to generate 3D scenes in one step},
    author={Wang, Hanyang and Liu, Fangfu and Chi, Jiawei and Duan, Yueqi},
    booktitle={2025 IEEE/CVF Conference on Computer Vision and Pattern Recognition (CVPR)},
    pages={16475--16485},
    year={2025},
    organization={IEEE}
}

@inproceedings{zhang2025world,
    title={World-consistent video diffusion with explicit 3d modeling},
    author={Zhang, Qihang and Zhai, Shuangfei and Martin, Miguel Angel Bautista and Miao, Kevin and Toshev, Alexander and Susskind, Joshua and Gu, Jiatao},
    booktitle={Proceedings of the Computer Vision and Pattern Recognition Conference},
    pages={21685--21695},
    year={2025}
}

@inproceedings{yan2025streetcrafter,
  title={Streetcrafter: Street view synthesis with controllable video diffusion models},
  author={Yan, Yunzhi and Xu, Zhen and Lin, Haotong and Jin, Haian and Guo, Haoyu and Wang, Yida and Zhan, Kun and Lang, Xianpeng and Bao, Hujun and Zhou, Xiaowei and others},
  booktitle={Proceedings of the Computer Vision and Pattern Recognition Conference},
  pages={822--832},
  year={2025}
}

@inproceedings{lin2023magic3d,
  title={Magic3D: High-Resolution Text-to-3D Content Creation},
  author={Lin, Chen-Hsuan and Gao, Jun and Tang, Luming and Takikawa, Towaki and Zeng, Xiaohui and Huang, Xun and Kreis, Karsten and Fidler, Sanja and Liu, Ming-Yu and Lin, Tsung-Yi},
  booktitle={IEEE Conference on Computer Vision and Pattern Recognition ({CVPR})},
  year={2023}
}

@InProceedings{chen2023fantasia3d,
  author={Chen, Rui and Chen, Yongwei and Jiao, Ningxin and Jia, Kui},
  title={Fantasia3D: Disentangling Geometry and Appearance for High-quality Text-to-3D Content Creation},
  booktitle={Proceedings of the IEEE/CVF International Conference on Computer Vision (ICCV)},
  year={2023},
  pages={22246-22256}
}

@article{zhang2024clay,
  title={Clay: A controllable large-scale generative model for creating high-quality 3d assets},
  author={Zhang, Longwen and Wang, Ziyu and Zhang, Qixuan and Qiu, Qiwei and Pang, Anqi and Jiang, Haoran and Yang, Wei and Xu, Lan and Yu, Jingyi},
  journal={ACM Transactions on Graphics (TOG)},
  volume={43},
  number={4},
  pages={1--20},
  year={2024},
  publisher={ACM New York, NY, USA}
}

@inproceedings{deitke2023objaverse,
    title={Objaverse: A universe of annotated 3d objects},
    author={Deitke, Matt and Schwenk, Dustin and Salvador, Jordi and Weihs, Luca and Michel, Oscar and VanderBilt, Eli and Schmidt, Ludwig and Ehsani, Kiana and Kembhavi, Aniruddha and Farhadi, Ali},
    booktitle={Proceedings of the IEEE/CVF conference on computer vision and pattern recognition},
    pages={13142--13153},
    year={2023}
}

@article{engstler2025syncity,
    title={Syncity: Training-free generation of 3d worlds},
    author={Engstler, Paul and Shtedritski, Aleksandar and Laina, Iro and Rupprecht, Christian and Vedaldi, Andrea},
    journal={arXiv preprint arXiv:2503.16420},
    year={2025}
}

@article{zheng2025constructing,
    title={Constructing a 3D Town from a Single Image},
    author={Zheng, Kaizhi and Zhang, Ruijian and Gu, Jing and Yang, Jie and Wang, Xin Eric},
    journal={arXiv preprint arXiv:2505.15765},
    year={2025}
}

@inproceedings{ho2021classifierfree,
    title={Classifier-Free Diffusion Guidance},
    author={Jonathan Ho and Tim Salimans},
    booktitle={NeurIPS 2021 Workshop on Deep Generative Models and Downstream Applications},
    year={2021},
}

@inproceedings{zhang2023adding,
  title={Adding conditional control to text-to-image diffusion models},
  author={Zhang, Lvmin and Rao, Anyi and Agrawala, Maneesh},
  booktitle={Proceedings of the IEEE/CVF international conference on computer vision},
  year={2023}
}

@article{ye2023ip-adapter,
  title={IP-Adapter: Text Compatible Image Prompt Adapter for Text-to-Image Diffusion Models},
  author={Ye, Hu and Zhang, Jun and Liu, Sibo and Han, Xiao and Yang, Wei},
  journal={arXiv preprint arxiv:2308.06721},
  year={2023}
}

@article{mou2023t2i,
  title={T2i-adapter: Learning adapters to dig out more controllable ability for text-to-image diffusion models},
  author={Mou, Chong and Wang, Xintao and Xie, Liangbin and Wu, Yanze and Zhang, Jian and Qi, Zhongang and Shan, Ying and Qie, Xiaohu},
  journal={arXiv preprint arXiv:2302.08453},
  year={2023}
}

@inproceedings{lipman2023flow,
  title={Flow Matching for Generative Modeling},
  author={Lipman, Yaron and Chen, Ricky TQ and Ben-Hamu, Heli and Nickel, Maximilian and Le, Matthew},
  booktitle={International Conference on Learning Representations},
  year={2023}
}

@article{saharia2022photorealistic,
  title={Photorealistic text-to-image diffusion models with deep language understanding},
  author={Saharia, Chitwan and Chan, William and Saxena, Saurabh and Li, Lala and Whang, Jay and Denton, Emily L and Ghasemipour, Kamyar and Gontijo Lopes, Raphael and Karagol Ayan, Burcu and Salimans, Tim and others},
  journal={Advances in neural information processing systems},
  volume={35},
  year={2022}
}

@inproceedings{rombach2022high,
  title={High-resolution image synthesis with latent diffusion models},
  author={Rombach, Robin and Blattmann, Andreas and Lorenz, Dominik and Esser, Patrick and Ommer, Bj{\"o}rn},
  booktitle={Proceedings of the IEEE/CVF conference on computer vision and pattern recognition},
  year={2022}
}

@article{bar2023multidiffusion,
  title={Multidiffusion: Fusing diffusion paths for controlled image generation},
  author={Bar-Tal, Omer and Yariv, Lior and Lipman, Yaron and Dekel, Tali},
  journal={International Conference on Machine Learning},
  year={2023}
}

@inproceedings{ding2023patched,
  title={Patched denoising diffusion models for high-resolution image synthesis},
  author={Ding, Zheng and Zhang, Mengqi and Wu, Jiajun and Tu, Zhuowen},
  booktitle={The twelfth international conference on learning representations},
  year={2023}
}

@inproceedings{jiang2025latent,
  title={Latent Patched Efficient Diffusion Model For High Resolution Image Synthesis},
  author={Jiang, Weiyun and Jangid, Devendra Kumar and Lee, Seok-Jun and Sheikh, Hamid Rahim},
  booktitle={Proceedings of the Computer Vision and Pattern Recognition Conference},
  year={2025}
}

@inproceedings{song2025progressive,
  title={Progressive Artwork Outpainting via Latent Diffusion Models},
  author={Song, Dae-Young and Yu, Jung-Jae and Cho, Donghyeon},
  booktitle={Proceedings of the IEEE/CVF International Conference on Computer Vision},
  year={2025}
}

@article{chen2024follow,
  title={Follow-your-canvas: Higher-resolution video outpainting with extensive content generation},
  author={Chen, Qihua and Ma, Yue and Wang, Hongfa and Yuan, Junkun and Zhao, Wenzhe and Tian, Qi and Wang, Hongmei and Min, Shaobo and Chen, Qifeng and Liu, Wei},
  journal={arXiv preprint arXiv:2409.01055},
  year={2024}
}

@article{wu2023panodiffusion,
  title={Panodiffusion: 360-degree panorama outpainting via diffusion},
  author={Wu, Tianhao and Zheng, Chuanxia and Cham, Tat-Jen},
  journal={arXiv preprint arXiv:2307.03177},
  year={2023}
}

@inproceedings{kim2021painting,
  title={Painting outside as inside: Edge guided image outpainting via bidirectional rearrangement with progressive step learning},
  author={Kim, Kyunghun and Yun, Yeohun and Kang, Keon-Woo and Kong, Kyeongbo and Lee, Siyeong and Kang, Suk-Ju},
  booktitle={Proceedings of the IEEE/CVF winter conference on applications of computer vision},
  year={2021}
}

@article{jimenez2023mixture,
  title={Mixture of diffusers for scene composition and high resolution image generation},
  author={Jim{\'e}nez, {\'A}lvaro Barbero},
  journal={arXiv preprint arXiv:2302.02412},
  year={2023}
}

@inproceedings{du2024demofusion,
  title={Demofusion: Democratising high-resolution image generation with no \$\$\$},
  author={Du, Ruoyi and Chang, Dongliang and Hospedales, Timothy and Song, Yi-Zhe and Ma, Zhanyu},
  booktitle={Proceedings of the IEEE/CVF conference on computer vision and pattern recognition},
  year={2024}
}

@article{lee2023syncdiffusion,
  title={Syncdiffusion: Coherent montage via synchronized joint diffusions},
  author={Lee, Yuseung and Kim, Kunho and Kim, Hyunjin and Sung, Minhyuk},
  journal={Advances in Neural Information Processing Systems},
  year={2023}
}

@article{lee2024streammultidiffusion,
  title={StreamMultiDiffusion: real-time interactive generation with region-based semantic control},
  author={Lee, Jaerin and Jung, Daniel Sungho and Lee, Kanggeon and Lee, Kyoung Mu},
  journal={Proceedings of the IEEE/CVF conference on computer vision and pattern recognition},
  year={2025}
}

@article{baek2025sonic,
  title={SONIC: Spectral Optimization of Noise for Inpainting with Consistency},
  author={Baek, Seungyeon and Dong, Erqun and Namazifard, Shadan and Matthews, Mark J and Yi, Kwang Moo},
  journal={arXiv preprint arXiv:2511.19985},
  year={2025}
}

@inproceedings{fu2024gptscore,
  title={Gptscore: Evaluate as you desire},
  author={Fu, Jinlan and Ng, See Kiong and Jiang, Zhengbao and Liu, Pengfei},
  booktitle={Proceedings of the 2024 Conference of the North American Chapter of the Association for Computational Linguistics: Human Language Technologies (Volume 1: Long Papers)},
  pages={6556--6576},
  year={2024}
}

@article{zhang2024gaussiancube,
  title={GaussianCube: A Structured and Explicit Radiance Representation for 3D Generative Modeling},
  author={Zhang, Bowen and Cheng, Yiji and Yang, Jiaolong and Wang, Chunyu and Zhao, Feng and Tang, Yansong and Chen, Dong and Guo, Baining},
  journal={NeurIPS},
  year={2024}
}
